\documentclass[twoside]{article}

\usepackage[accepted]{aistats2025}
\usepackage[round]{natbib}

\usepackage{nicefrac}
\usepackage{amssymb,latexsym,amsfonts,amsmath,amsthm}
\usepackage{color}
\usepackage{enumerate}
\usepackage{enumitem}
\usepackage{algorithm}
\usepackage{algpseudocode}
\usepackage{bm}
\usepackage{subfigure} 
\usepackage{mathtools}
\usepackage[titletoc,title]{appendix}
\usepackage[T1]{fontenc}
\usepackage{dsfont}
\usepackage{url}
\usepackage[font=small,labelfont=bf]{caption}
\usepackage{hyperref}[] 
\usepackage{cleveref}
\usepackage{thm-restate}
\usepackage{placeins}

\newcommand{\mP}{\mathbb{P}}
\newcommand{\mE}{\mathbb{E}}

\newcommand{\tXn}{{\widetilde{\mathcal{X}}_{n}}}
\newcommand{\Xnm}{{\mathcal{X}_{n+m}}}
\newcommand{\tXnm}{{\widetilde{\mathcal{X}}_{n+m}}}

\newcommand{\bpi}{\boldsymbol{\pi}}

\newcommand{\given}{\,|\,} 

\newcommand{\etc}{\emph{etc}}

\newtheorem{assumption}{Assumption}

\begin{document}





\twocolumn[

\aistatstitle{Robust Kernel Hypothesis Testing under Data Corruption}

\aistatsauthor{ 
Antonin Schrab$^\star$
\And 
Ilmun Kim$^\star$

}

\aistatsaddress{
Centre for Artificial Intelligence \\ 
Gatsby Computational Neuroscience Unit \\
University College London \& Inria London
\And  
Department of Statistics and Data Science \\ 
Department of Applied Statistics \\
Yonsei University
} 
]


\begin{abstract}
We propose a general method for constructing robust permutation tests under data corruption. The proposed tests effectively control the non-asymptotic type I error under data corruption, and we prove their consistency in power under minimal conditions. This contributes to the practical deployment of hypothesis tests for real-world applications with potential adversarial attacks. For the two-sample and independence settings, we show that our kernel robust tests are minimax optimal, in the sense that they are guaranteed to be non-asymptotically powerful against alternatives uniformly separated from the null in the kernel MMD and HSIC metrics at some optimal rate (tight with matching lower bound). We point out that existing differentially private tests can be adapted to be robust to data corruption, and we demonstrate in experiments that our proposed tests achieve much higher power than these private tests. Finally, we provide publicly available implementations and empirically illustrate the practicality of our robust tests. 
\end{abstract}

\section{INTRODUCTION}
\label{sec:introduction}

One of the fundamental goals of statistical machine learning is to quantify departures from non-parametric distributional null hypotheses, with various applications in causal discovery, model calibration, clinical trials, \etc. 
However, after collecting real-world data, practitioners rarely observe that the null holds. 
This can be explained by the inherently intricate process that data collection represents. 
In fact, \emph{real data is messy}, often consisting of outliers and corrupted or wrong values. 
This illustrates that these nulls are often too strong to be observed in real-world applications, hence ultimately preventing practitioners from using such tests.
To overcome these practical challenges, we consider a framework with relaxed null hypotheses for which distributional conditions only need to hold on a large proportion of the data (as opposed to the entire dataset), hence facilitating the deployment of robust tests in practice. Specifically, we focus on scenarios where some data points are arbitrarily corrupted by an adversary, and aim to develop methods for constructing robust tests that fulfill the following desiderata: the considered tests are (i) straightforward to implement, (ii) applicable to an important set of nonparametric testing problems, (iii) maintain non-asymptotic validity under data corruption, and (iv) attain minimax optimal power in certain settings. With this goal in mind, we first review some relevant work and highlight our key contributions.

\textbf{Robust testing.} There is an extensive body of work on robust testing developed under various contamination models. One popular framework is Huber's $\epsilon$-contamination model~
\citep{huber1964} 
where the observed data are assumed to be drawn from a mixture distribution $(1-\epsilon) P + \epsilon G$ with $P$ as the target distribution and $G$ as the contamination distribution. Our work is concerned with a stronger adversarial contamination model where an adversary selects $r$ observations out of $n$ data points and replace them with arbitrary values. Under different contamination models, there has been a flurry of recent work on robust testing such as the two-point testing problem~\citep{chen2016general,li2023robustness}, covariance testing~\citep{diakonikolas2021sample}, mean testing~\citep{diakonikolas2017,george2022robust,canonne2023}, identity testing~\citep{acharya2021robust}. See \citet{diakonikolas2019recent} for a recent survey on robust statistics. We aim to advance the field by studying robust testing for exchangeability (see \Cref{subsec:framework} for a formal setup), and establishing minimax testing rates in terms of the kernel metrics under data corruption. 
A distinct line of work explores a different notion of robustness, where the goal is to determine whether a given dataset lies within an uncertainty set centered around an empirical distribution with uncertainty defined using various metrics~\citep{gao2018robust,wang2022data,sun2021data,sun2022robust,sun2023kernel}. This approach contrasts with our focus on worst-case contamination as it seeks to account for distributional ambiguity rather than adversarial data corruption.

\textbf{Permutation tests.} As mentioned earlier, our main interest is in designing robust tests for assessing the exchangeability of data under the null hypothesis. A gold standard method for testing exchangeability is the permutation test, which leverages the permutation-invariant properties of the data distribution under the null hypothesis. This ensures rigorous control of the type I error rate as demonstrated in \citet[\emph{e.g.},][]{romano2005exact,hemerik2018exact}. When the data are arbitrarily corrupted, however, the exchangeability assumption no longer holds, and naively applying the permutation test using the corrupted data can potentially inflate the type I error rate. To address this miscalibration issue, it is essential to either modify the test statistic or adjust the permutation critical value under data corruption. This process inevitably compromises the power of the test, and the key challenge is then to balance a trade-off between robustness and power. We approach this problem under the minimax testing framework as in \citet{albert2019adaptive,schrab2021mmd,acharya2021robust,kim2023differentially} and explain the fundamental limit of testing under data corruption.

\textbf{Kernel MMD and HSIC.} 
The Maximum Mean Discrepancy (MMD, \citealp{gretton2012kernel}) and the Hilbert--Schmidt Independence Criterion (HSIC,  \citealp{gretton2005measuring}) are two prominent kernel-based measures for assessing homogeneity and dependence between two random quantities. Since their introduction, there have been significant developments in related topics, including time-efficient kernel tests~\citep{gretton2012optimal,zaremba2013b,zhao2015fastmmd,yamada2019post,schrab2022efficient,domingo2023compress}, adaptive kernel selections~\citep{schrab2021mmd,schrab2022efficient,schrab2022ksd,hagrass2022spectral,domingo2023compress,biggs2023mmdfuse,chatterjee2023boosting} and minimax optimality of kernel testing~\citep{albert2019adaptive,li2019optimality,schrab2021mmd,shekhar2022two,shekhar2022ind,hagrass2022spectral}. However, prior work has predominantly focused on standard i.i.d.~data without any perturbation, potentially limiting their applicability to real-world scenarios with anomalies or adversarial attacks. 
While there exist some studies on the robustness of MMD statistics in terms of estimation (\citealp[Section 3.3]{briol2019statistical} and \citealp[Section 3.2.3]{cherief2022finite}) and in terms of Bayesian inference \citep{cherief2020mmd,dellaporta2022robust,dellaporta2023robust,legramanti2022concentration,fazeli2024semi}, there has been limited understanding of their performance in robust testing.
To fill this gap, we consider the two-sample and independence testing problems as running examples of our general methods, and propose robust tests based on the MMD and HSIC statistics. We then show that the proposed  tests guarantee rigorous control of testing errors even in the presence of data corruption, and achieve minimax optimality across all levels of data corruption.

\textbf{Outline and summary.}
The remainder of this paper is organised as follows. In \Cref{sec:datacorruption}, we formally introduce our \emph{testing under data corruption} framework, and propose our DC procedure (\Cref{alg:dc}) for constructing robust permutation tests from any statistic with finite sensitivity. We prove that the validity of this procedure (non-asymptotic control of the type I error even under data corruption) and present conditions for consistency (asymptotic control of the type II error).
In \Cref{sec:twosample,sec:independence}, we construct robust kernel tests for the two-sample and independence testing frameworks, respectively. For each robust test, we prove minimax optimality under data corruption in terms of the respective MMD and HSIC kernel metrics. 
In \Cref{sec:related}, we discuss related work on differential privacy which can be leveraged to construct robust tests.
We empirically verify the validity of all robust tests in \Cref{sec:experiments}, and highlight, on corrupted synthetic and real-world data, the significantly higher power achieved by our DC tests compared to alternative private tests of \citet{kim2023differentially}.
\Cref{sec:conclusion} closes the paper with further discussions. 
We defer the technical proofs of the main results to the appendix.

\textbf{Notation.}
Consider two ordered sets $\mathcal{X}_n \coloneqq (X_1,\ldots,X_{n})$ and ${\mathcal{Y}}_n\coloneqq ( Y_1,\ldots, Y_{n})$.
The Hamming distance $d_{\mathrm{ham}}(\mathcal{X}_n, {\mathcal{Y}}_n) := \sum_{i=1}^n \mathds{1}(X_i \neq Y_i)$ is the number of indices for which the entries of the two ordered sets are different.
We let $[n]$ denote the set $\{1,\dots,n\}$ and $[n]_0\coloneqq\{0,\dots,n\}$. 
We define $\boldsymbol{\Pi}_n$ to be the set of all permutations of $[n]$. 
Given a permutation $\bpi\in\boldsymbol{\Pi}_n$, we denote by $\mathcal{X}_n^{\bpi}$ the permuted ordered set $(X_{\bpi(1)},\ldots,X_{\bpi(n)})$. For two sequences of real numbers $a_n$ and $b_n$, we write $a_n \lesssim b_n$ if $a_n \leq C b_n$ for some constant $C>0$, and write $a_n \asymp b_n$ if $a_n \lesssim b_n$ and $b_n \lesssim a_n$. The global sensitivity $\Delta_T$ of a statistic $T$ is defined as the maximum difference in the statistic's output when evaluated on permuted datasets whose entries differ by at most one, that is
\begin{equation*}
\Delta_T \coloneqq \sup_{\bpi \in \boldsymbol{\Pi}_n} \sup_{\substack{\mathcal{X}_{n},{\mathcal{Y}}_{n}\,:\,d_{\mathrm{ham}}(\mathcal{X}_{n},{\mathcal{Y}}_{n}) \leq 1}} \bigl| T(\mathcal{X}_{n}^{\bpi}) - T({\mathcal{Y}}_{n}^{\bpi}) \bigr|.
\end{equation*} 
Given values $M_0,\ldots,M_B$, the $(1\!\!-\!\!\alpha)$-quantile of the $M_i$'s is defined as
$q_{1-\alpha} \coloneqq \inf \bigl\{ t \in \mathbb{R} : \frac{1}{B+1} \sum_{i=0}^{B} \mathds{1}(M_i \leq t) \geq 1- \alpha  \bigr\}$. 
We let $\mathsf{Laplace}(0,1)$ denote a Laplace distribution with location and scale parameters $(0,1)$. 
We also let $\mathsf{Gaussian}(\mu,\sigma,d)$ denote a $d$-dimensional distribution with each dimension being independent and following a Gaussian distribution with mean $\mu$ and standard deviation $\sigma$.
Similarly, $\mathsf{Geometric}(p,d)$ denotes a $d$-dimensional distribution with each dimension being independent and following a geometric distribution taking values in $\{0,1,2,\ldots\}$ with parameter $p\in(0,1)$.

\section{CONSTRUCTION OF HYPOTHESIS TESTS ROBUST TO DATA CORRUPTION}
\label{sec:datacorruption}

In \Cref{subsec:framework}, we set the stage and formalise our framework for robust testing in the present of  data corruption.
In \Cref{subsec:dctest}, we introduce a novel procedure (DC) that enables the construction of robust tests under data corruption using any test statistic with finite sensitivity.

\subsection{Robust testing framework under data corruption}
\label{subsec:framework}

\textbf{Standard testing framework.}
Consider a set of distributions $\mathcal{P}$ partitioned into disjoint subsets $\mathcal{P}_0$ and $\mathcal{P}_1$.
Given a set $\widetilde{\mathcal{X}}_n$ consisting of $n$ random samples drawn i.i.d.~from $P \in \mathcal{P}$, the aim of hypothesis testing is to test whether the null $\mathcal{H}_0\colon P \in \mathcal{P}_0$, or the alternative $\mathcal{H}_1\colon P \in \mathcal{P}_1$, holds.

\textbf{Testing under data corruption framework.}
In this setting, we do not have access to $\widetilde{\mathcal{X}}_n$ but only to a corrupted version of it, denoted simply by $\mathcal{X}_n$, where up to $r$ samples of $\widetilde{\mathcal{X}}_n$ might have been corrupted (possibly in an adversarial manner). We receive only the set $\mathcal{X}_n$ of size $n$ with no knowledge of which samples, if any, have been corrupted. The aim is still to test whether $\mathcal{H}_0\colon P \in \mathcal{P}_0$ or $\mathcal{H}_1\colon P \in \mathcal{P}_1$, but we can only assume $n-r$ samples of $\mathcal{X}_n$ are actually drawn from $P \in \mathcal{P}$ (as $r$ samples might have been manipulated).

\textbf{Robustness.}
Hypothesis tests designed for this setting are \emph{robust} to data corruption: under the null, manipulating up to $r$ samples will not make the test deviate from the null.
This can also be thought of as enlarging the null hypothesis as it only needs to hold for at least $n-r$ samples of the data rather than for all $n$ samples.
The number $r$ of maximum sample manipulations to be robust to is specified by the user depending on the application.
If $r=0$, we recover the standard testing framework.
As $r$ increases, the test becomes more robust but less powerful (\emph{i.e.}, there is a trade-off between robustness and power). If $r=n$, the null would never be rejected.

\textbf{Exchangeability.}
We restrict our attention to testing frameworks under which the exchangeability assumption holds \citep[][Chapter 15.2]{lehmann2005testing} in the non-corrupted setting, which means that, under the null, $\widetilde{\mathcal{X}}_n$ is exchangeable: for any permutation $\bpi \in \boldsymbol{\Pi}_n$, the joint distributions of $\widetilde{\mathcal{X}}_n$, and of $\widetilde{\mathcal{X}}_n^{\bpi}$, are the same. For example, the two-sample and independence testing frameworks, serving as primary applications of our proposal, satisfy this assumption.

\subsection{DC procedure: Robust test construction against data corruption}
\label{subsec:dctest}

We assume the maximum number of corrupted samples $r$ to be fixed apriori.
The aim is then to construct a robust test that is \emph{valid} in the sense that it controls the type I error at the desired level $\alpha$ even when up to $r$ samples are arbitrarily corrupted.

We now propose a procedure to construct tests which are robust to data corruption.
First, recall that, given a statistic $T$, the classical permutation test is defined as rejecting the null when $T_0 > q_{1-\alpha}(T_0\ldots,T_B)$ where $T_0 = T(\mathcal{X}_n)$ and $T_i = T(\mathcal{X}_n^{\bpi_i})$, $i\in[B]$ for $B$ permutations $\bpi_1,\ldots,\bpi_B\in\boldsymbol{\Pi}_n$. 
Such a test in the usual setting (no data corruption) controls the type I error at level $\alpha$ non-asymptotically~\citep[\emph{e.g.},][]{hemerik2018exact}.
For our setting with at most $r$ corrupted samples, we instead define our DC test as rejecting the null when $T_0 > q_{1-\alpha}(T_0\ldots,T_B) + 2 r \Delta_T$ (see details in \Cref{alg:dc}), where $\Delta_T$ is the global sensitivity of $T$.
We prove that this results in a well-calibrated non-asymptotic test under $r$ data corruption, and provide sufficient conditions to guarantee its consistency.

\vspace{0.5cm}

\begin{minipage}{0.49\textwidth}
	\begin{algorithm}[H]\raggedright \caption{Robust DC procedure} 
		\label{alg:dc}
		\textbf{Inputs:} Data $\mathcal{X}_n$, robustness $r$, level $\alpha$, statistic $T$, permutation number $B$.\\
	\vskip 1.05em
		Generate i.i.d. permutations $\bpi_1,\dots,\bpi_B$ of $[n]$. \\
		Set $\bpi_0=\mathrm{Id}$ and compute global sensitivity $\Delta_T$. \\
		Compute $T_i = T(\mathcal{X}_n^{\bpi_i})$, $i\in[B]_0$. \\
		Compute $(1\!-\!\alpha)$-quantile $q$ of $T_0,\dots,T_B$. \\
	\vskip 1.05em
	\textbf{Output}: Reject $\mathcal{H}_0$ if $T_0 > q+ 2 r \Delta_T$.
\end{algorithm}
\end{minipage}

\vspace{1.3cm}

\begin{restatable}[Validity \& consistency of DC]{lemma}{dcconsistent}
\label{res:dc_consistent}
(i) The DC test of \Cref{alg:dc} has non-asymptotic level control under $r$ data corruption.
(ii) Let $P_1\in\mathcal{P}_1$ be an alternative distribution. Assume $\alpha\in(0,1)$ fixed.
For any sequence of $B_n$ of permutation numbers satisfying $\min_{n\in\mathbb{N}} B_n > \alpha^{-1} - 1$, the DC test is consistent in the sense that $\lim_{n \rightarrow \infty} \mP_{P_1}\bigl(\textrm{\emph{DC rejects $\mathcal{H}_0$}} \given r \textrm{\emph{ corrupted data}} \bigr) = 1$ if
$$
\lim_{n \rightarrow \infty} \mP_{P_1}\bigl(\,T(\tXn) ~>~ T(\tXn^{\bpi}) + 4 r\Delta_T \bigr) = 1,
$$
where the probability is taken with respect to the (uniformly) random permutation $\bpi$ of $[n]$, and to $\tXn$ i.i.d.~drawn from $P_1$.
\end{restatable}

We highlight that \Cref{res:dc_consistent} is applicable to any test statistic with finite global sensitivity $\Delta_T$. 
Moreover, the validity under data corruption holds in any testing framework for which exchangeability holds under the (uncorrupted) null hypothesis. 
Finally, we stress that knowledge of the maximum number of potential corruptions $r$ is sufficient for all theoretical guarantees in this paper, including \Cref{res:dc_consistent}, to hold; knowing the exact number of corrupted samples is not required.

We next turn to two-sample and independence kernel testing as specific applications of these general methods, and present more refined results.

\section{TWO-SAMPLE KERNEL TESTING ROBUST TO DATA CORRUPTION}
\label{sec:twosample}

\textbf{Robust two-sample testing.}
We now consider the two-sample testing framework which, in the non-corrupted setting, satisfies the exchangeability assumption under the null \citep[\emph{e.g.},][Proposition 1]{schrab2021mmd}.
In our robust setting, we are given samples $Y_1,\ldots,Y_n$  and $Z_1,\ldots,Z_m$, where at most $r$ of the $m+n$ samples have been corrupted (we do not know which ones, if any). Of the remaining samples (at least $m+n-r$ of them), the $Y_i$'s are drawn from $P$ while the $Z_i$'s are drawn from $Q$. We are interested in testing whether $P=Q$.
The data corruption could potentially be applied in an adversarial way; for example, some of the $Y_i$'s samples could be replaced by new samples from $Q$.
As before, the maximum number of corrupted samples $r$ is specified by the user and is considered to be known.
We restrict ourselves to the case $r \leq n = \min(m,n)$, as the setting $r>n$ suffers from the same issue as when $r = n$: all the samples from $P$ could be corrupted, in which case all information about $P$ would be lost and we would not be able to test whether $P=Q$.

\textbf{MMD statistic.}
As a divergence for this two-sample problem, we use the kernel MMD introduced by \citet{gretton2012kernel} which, for a kernel $k$, is defined as
\begin{align*}
	&\mathrm{MMD}_k(P,Q) \\ 
	\coloneqq ~&\sqrt{\mE_P[k(Y,Y')]  - 2\mE_{P,Q}[k(Y,Z)] + \mE_Q[k(Z,Z')]} 
\end{align*}
for probability distributions $P$ and $Q$.
The MMD is well-suited for the two-sample problem as it can distinguish between any two distributions in the sense that $\mathrm{MMD}_k(P,Q) = 0$ if and only if $P=Q$, given that the kernel $k$ is characteristic \citep{fukumizu2008kernel}.
For a given sample $\Xnm \coloneqq (Y_1,\ldots,Y_n,Z_1,\ldots,Z_m)$ generated as described above, the quadratic-time empirical MMD statistic is
\begin{align*}
	&\widehat{\mathrm{MMD}}(\mathcal{X}_{n+m})\\
	\coloneqq ~&\bigg(\frac{1}{n^2} \sum_{i,j=1}^{n} k(Y_i,Y_j) +  \frac{1}{m^2} \sum_{i,j=1}^{m} k(Z_i,Z_j) \\
		  &\phantom{\bigg(\frac{1}{n^2} \sum_{i,j=1}^{n} k(Y_i,Y_j) }
  - \frac{2}{nm} \sum_{i=1}^n\sum_{j=1}^m k(Y_i,Z_j)\bigg)^{1/2}
\end{align*}
which is a plug-in estimator, and its square can be expressed as a two-sample V-statistic. For a kernel bounded everywhere by $K$, \citet[Lemma 5]{kim2023differentially} provide an upper bound on the global sensitivity of this MMD statistic 
$\Delta_{\widehat{\mathrm{MMD}}} \leq \sqrt{2K}/ \min(n,m)$, which is guaranteed to be tight when the kernel $k$ is translation invariant. 

\textbf{dcMMD.}
We construct our robust two-sample dcMMD test by applying the procedure of \Cref{alg:dc} with the MMD statistic $T=\widehat{\mathrm{MMD}}$ and global sensitivity $\Delta_{T} = \sqrt{2K}/\min(n,m)$. 
It is immediately clear from \Cref{res:dc_consistent} that the robust dcMMD test controls the type I error at the desired level $\alpha$ non-asymptotically, even under $r$ data corruption.
We next prove its consistency in the data corruption framework, which ensures that any fixed distributions $P$ and $Q$ with $P\neq Q$ can be distinguished with probability one by the dcMMD test for large enough sample sizes. 
This test scales quadratically with the sample sizes.

\begin{restatable}[Consistency of dcMMD]{lemma}{mmdconsistent} 
	\label{res:mmd_consistent}
    Suppose that the kernel $k$ is characteristic, non-negative, and bounded everywhere by $K$. 
    Assume that $n \leq m$ and $r/n \to 0$ as $n\to\infty$, and that a sequence of permutation numbers $B_n$ satisfies $\min_{n\in\mathbb{N}} B_n > \alpha^{-1} - 1$. Then, for any fixed $P$ and $Q$ with $P \neq Q$, dcMMD is consistent in the sense that 
    $$
    \lim_{n \rightarrow \infty} \mP_{P,Q}\bigl(\textrm{\emph{reject $\mathcal{H}_0$}} \given r \textrm{\emph{ corrupted data}} \bigr) = 1.
	$$
\end{restatable}

The proof of \Cref{res:mmd_consistent} relies on proving that the sufficient condition of \Cref{res:dc_consistent} is satisfied. We show that this condition is met for the proposed dcMMD test with characteristic and bounded kernel. Having obtained guarantees for the asymptotic power of dcMMD against fixed alternatives, we now consider the more challenging task of providing power guarantees that hold uniformly over classes of alternatives shrinking towards the null as the sample sizes increase.
We choose these classes to be separated from the null in the MMD metric, and prove that dcMMD achieves the optimal uniform separation rate (with matching upper and lower bounds).
This minimax result holds with respect to the smallest sample size, to the data corruption level, and to the quantities controlling the type I and II errors; hence, demonstrating the optimality of dcMMD in the data corruption setting.

\begin{restatable}[Minimax optimal uniform separation of dcMMD]{theorem}{mmdseparation}  \label{res:mmd_separation}
	Suppose that the kernel $k$ is characteristic, non-negative, and bounded everywhere by $K$. For $\alpha,\beta \in (0,1)$, assume that the number of permutations $B$ is greater than $3 \alpha^{-2}\bigl\{ \log\bigl(8/\beta\bigr) + \alpha (1-\alpha) \}$%
, and that $n \asymp m$ with $n\leq m$.
	\begin{enumerate}[leftmargin=*,nolistsep]
		\item[(i)] (Uniform separation) The dcMMD test is guaranteed to have high power, i.e.,
		$\mP_{P,Q}\bigl(\textrm{\emph{reject $\mathcal{H}_0$}} \given r \textrm{\emph{ corrupted data}} \bigr) \geq 1 -\beta$
		for any distributions $P$ and $Q$ separated as
		\begin{align*}
			&\mathrm{MMD}_k(P,Q)\\
			\geq~
			&C_{K} \max \biggl\{ \sqrt{\frac{\max\{\log(e/\alpha),\log(e/\beta)\}}{n}}, \ \frac{r}{n} \biggr\} 
		\end{align*}
		for some positive constant $C_{K}$ depending on $K$.\\[-0.6em]
		\item[(ii)] (Minimax optimality) Further assuming that the kernel is translation invariant and non-constant as in \Cref{assumption: two-sample kernel} in the appendix, this separation rate is optimal in terms of the smallest sample size $n$, of the data corruption level $r$, and of the testing errors $\alpha,\beta$ with $\alpha \asymp \beta$ and $\alpha+\beta <0.4$.
	\end{enumerate}
\end{restatable}

\Cref{res:mmd_separation} demonstrates that the dcMMD test achieves minimax rate optimality in terms of the MMD metric. 
Further remarks regarding \Cref{res:mmd_separation} are in order, which we present below.

\textbf{Low corruption regime.} We first note that when $r \lesssim \sqrt{n\max\{\log(e/\alpha),\log(e/\beta)\}}$, the dominating term in the condition of \Cref{res:mmd_separation}(i) is the first one, which is the minimax rate in the usual no-corruption setting~\citep[][Theorem 8]{kim2023differentially}. 
In this low corruption regime, the two robust dcMMD tests achieve the same minimax optimal rate as the standard MMD test in the non-corrupted setting.
This demonstrates that we can obtain robustness against corruption of up to $r$ samples, where $r \lesssim \sqrt{n\max\{\log(e/\alpha),\log(e/\beta)\}}$, without compromising 
power in terms of uniform separation rate.

\textbf{High corruption regime.} 
When we have $\sqrt{n\max\{\log(e/\alpha),\log(e/\beta)\}} \lesssim r < n$, our robust dcMMD test achieves high power when $\mathrm{MMD}_k$ exceeds $r/n$, which is the optimal rate in this high corruption regime as we have proved in \Cref{res:mmd_separation}(ii). It is worth pointing out that the separation rate $r/n$ in the high corruption regime is independent of $\alpha$ and $\beta$, unlike the one obtained in the low corruption regime. However, we emphasise that both $\alpha$ and $\beta$ cannot be taken to be arbitrarily small as this would break the condition $\sqrt{n\max\{\log(e/\alpha),\log(e/\beta)\}} \lesssim r$.

\textbf{Total corruption regime.} 
When $r= n$ (or $r\geq n$), the separation rate becomes an arbitrary constant. This makes the condition for $\mathrm{MMD}_k$ in \Cref{res:mmd_separation}(i) vacuous since $\mathrm{MMD}_k$ is bounded above by $\sqrt{2K}$ for any distributions $P$ and $Q$. In fact, in this regime, all samples from $P$ could be adversarially replaced by samples from $Q$, so there is no hope of distinguishing between the two distributions. 

\section{INDEPENDENCE KERNEL TESTING ROBUST TO DATA CORRUPTION}
\label{sec:independence}

\textbf{Robust independence testing.}
Another testing framework satisfying the exchangeability assumption under the null is the independence testing one \citep[\emph{e.g.},][Proposition 1]{albert2019adaptive}.
The robust independence testing problem is defined as follows.
Given paired samples $\mathcal{X}_n = \big((Y_i,Z_i)\big)_{i=1}^n$ where at most $r\in[n]$ of them have been corrupted (no knowledge of which ones are corrupted, if any), and the remaining paired samples (at least $n-r$ of them) are drawn from some joint distribution $P_{Y\!Z}$.
We test whether that joint distribution $P_{Y\!Z}$ is equal to the product of its marginals $P_Y\times P_Z$, that is, we test for independence among the non-corrupted paired samples.
Again, the maximum number of corrupted samples $r$ is considered known, usually specified by the user.

\textbf{HSIC statistic.}
As a measure of dependence, we use the HSIC~\citep{gretton2005kernel} which, for kernels $k$ and $\ell$, is defined as
\begin{align*}
	\mathrm{HSIC}_{k,\ell}(P_{Y\!Z}) \coloneqq \biggl(&\mE_{P_{Y\!Z}}\!\big[k_{Y,Y'}\ell_{Z,Z'}\big] \\
	&- 2 \mE_{P_{Y\!Z}}\!\big[\mE_{P_{Y}}[k_{Y,Y'}]\mE_{P_{Z}}[\ell_{Z,Z'}]\big] \\ 
	&+\mE_{P_Y}\!\big[k_{Y,Y'}\big] \mE_{P_Z}\!\big[\ell_{Z,Z'}\big]\biggr)^{1/2}
\end{align*}
with the condensed notation $k_{Y,Y'}$ and $\ell_{Z,Z'}$ for $k(Y,Y')$ and $\ell(Z,Z')$.
Provided that both kernels are characteristic \citep{gretton2015simpler}, the HSIC characterises dependence in the sense that $\mathrm{HSIC}_{k,\ell}(P_{Y\!Z}) = 0 $ if and only if $P_{Y\!Z} = P_Y \times P_Z$; hence, it is suitable as a building block for independence testing. 
Given paired samples $\mathcal{X}_n = \big((Y_i,Z_i)\big)_{i=1}^n$ generated by the process described above, the empirical HSIC statistic is
\begin{align*} 
	\widehat{\mathrm{HSIC}}(\mathcal{X}_n) ~\coloneqq~  \bigg(&\frac{1}{n^2} \sum_{i,j=1}^n k_{i,j} \ell_{i,j} - \frac{2}{n^3} \sum_{i,j_1,j_2=1}^n k_{i,j_1} \ell_{i,j_2}	\\
	&+ \frac{1}{n^4} \sum_{i_1,i_2,j_1,j_2=1}^n k_{i_1,j_1} \ell_{i_2,j_2} \bigg)^{1/2}
\end{align*}
where $k_{i,j}$ and $\ell_{i,j}$ denote $k(Y_i,Y_j)$ and $\ell(Z_i,Z_j)$ for $i,j\in[n]$.
This quadratic-time plug-in estimator can be written as a one-sample fourth-order V-statistic.
For kernels $k$ and $\ell$ bounded everywhere by $K$ and $L$, respectively, the global sensitivity of the HSIC statistic is bounded as
$\Delta_{\widehat{\mathrm{HSIC}}} \leq 4 \sqrt{KL}(n-1)/n^2$, which is asymptotically tight as shown by \citet[Lemma 6]{kim2023differentially}. 

\textbf{dcHSIC.}
Under the data corruption setting, the robust dcHSIC is constructed by applying the DC procedure of \Cref{alg:dc} with the HSIC statistic $T=\widehat{\mathrm{HSIC}}$ and global sensitivity $\Delta_{T}$ as $4 \sqrt{KL}(n-1)/n^2$.
The validity of dcHSIC is guaranteed by \Cref{res:dc_consistent} with non-asymptotic type I error control at the desired level $\alpha$ under $r$ data corruption. 
The next lemma proves that the dcHSIC test is consistent against any fixed alternative $P_{Y\!Z}$ with $P_{Y\!Z} \neq P_Y\times P_Z$. 
In other words, this test can detect any fixed dependence between $Y$ and $Z$ with probability one for large enough sample size.
The dcHSIC test scales quadratically with $n$.

\begin{restatable}[Consistency of dcHSIC]{lemma}{hsicconsistent} 
	\label{res:hsic_consistent}
	Suppose that the kernels $k$ and $\ell$ are characteristic, non-negative, and bounded everywhere by $K$ and $L$, respectively. 
    Assume that $r/n \to 0$ as $n \to \infty$, and that a sequence of permutation numbers $B_n$ satisfies $\min_{n\in\mathbb{N}} B_n > \alpha^{-1} - 1$. Then, for any fixed joint distribution $P_{Y\!Z}$ with $P_{Y\!Z}\neq P_Y \times P_Z$, dcHSIC is consistent in the sense that 
    $$
    \lim_{n \rightarrow \infty} \mP_{P_{Y\!Z}}\bigl(\textrm{\emph{reject $\mathcal{H}_0$}} \given r \textrm{\emph{ corrupted data}} \bigr) = 1.
    $$
\end{restatable}

Similarly to the robust two-sample testing setting, we can also obtain power guarantees for dcHSIC in terms of uniform separation rates.
More precisely, the type II errors of dcHSIC can be controlled non-asymptotically whenever $\mathrm{HSIC}_{k,\ell}(P_{Y\!Z})$ is larger than the minimax rate (which tends to zero as the sample size grows), highlighting the optimality of dcHSIC under the data corruption framework.

\begin{restatable}[Minimax optimal uniform separation of dcHSIC]{theorem}{hsicseparation} 
	\label{res:hsic_separation}
	Suppose that the kernels $k$ and $\ell$ are characteristic, non-negative, and bounded everywhere by $K$ and $L$, respectively. For $\alpha,\beta \in (0,1)$, assume that the number of permutations $B$ is greater than $\alpha^{-2}\bigl\{ \log\bigl(8/\beta\bigr) + \alpha (1-\alpha) \}$.
	\begin{enumerate}[leftmargin=*,nolistsep]
		\item[(i)] (Uniform separation) The dcHSIC test is guaranteed to have high power, i.e.,
		$\mP_{P_{Y\!Z}}\bigl(\textrm{\emph{reject $\mathcal{H}_0$}} \given r \textrm{\emph{ corrupted data}} \bigr) \geq 1 -\beta$ for any joint distribution $P_{Y\!Z}$ separated as
		\begin{align*}
			&\mathrm{HSIC}_{k,\ell}(P_{Y\!Z})
			\\
			~\geq~&C_{K,L} \max \biggl\{ \sqrt{\frac{\max\{\log(e/\alpha),\log(e/\beta)\}}{n}}, \ \frac{r}{n} \biggr\} 
		\end{align*}
		for some positive constant $C_{K,L}$ depending on $K$ and $L$. \\[-0.6em]
		\item[(ii)] (Minimax optimality) Further assuming that the kernels $k$ and $\ell$ are translation invariant and non-constant as in \Cref{assumption: independence kernel} in the appendix, this separation rate is optimal in terms of the sample size $n$, of the data corruption level $r$, and of the testing errors $\alpha,\beta$ with $\alpha \asymp \beta$ and $\alpha+\beta <0.4$.
	\end{enumerate}
\end{restatable}

The interpretation of the independence uniform separation rate of \Cref{res:hsic_separation} mirrors the one following \Cref{res:mmd_separation} for the two-sample case.
In the low corruption regime with $r \lesssim \sqrt{n\max\{\log(e/\alpha),\log(e/\beta)\}}$, robustness comes for free as the minimax rate is the same as in the non-corrupted setting. 
In the high corruption regime with $\sqrt{n\max\{\log(e/\alpha),\log(e/\beta)\}} \lesssim r < n$, only the alternatives for which $\mathrm{HSIC}_{k,\ell}(P_{Y\!Z})$ exceeds $r/n$ can be detected with high power, and this rate is optimal with respect to both $n$ and $r$ by \Cref{res:hsic_separation}(ii). In the total corruption regime where $r=n$, the uniform separation condition is never satisfied and any test is non-informative as all information is lost. Finally, we highlight again that our dcHSIC test is minimax optimal across all regimes of data corruption.


\section{RELATED WORK: ROBUSTNESS VIA DIFFERENTIAL PRIVACY}
\label{sec:related}







Robust tests can also be constructed from existing differentially private tests.
In \Cref{subsec:dp_background}, we present background information on differential privacy.
In \Cref{subsec:dptest}, we construct robust tests by leveraging the existing differential privatisation (DP) method  of \citet{kim2023differentially} for permutation tests. We prove their validity and defer its power analysis to \Cref{sec:dp_results}.

\subsection{Differential privacy background}
\label{subsec:dp_background}

Differential privacy~\citep{dwork2006calibrating} is a popular framework for private data analysis that aims to return an output insensitive to any single individual's data point. As discussed in \citet{dwork2009differential,avella2020}, there is a close connection between robust statistics and differential privacy. Both frameworks require that algorithms remain stable under data perturbation. Recent research has delved into extending non-private tests to privacy settings and studying their asymptotic properties~\citep{fienberg2011privacy,couch2019differentially,campbell2018differentially,rogers2017new,raj2020differentially} and non-asymptotic properties~\citep{cai2017priv,aliakbarpour2017differentially,aliakbarpour2019private,acharya2019test,kim2023differentially}. 

\subsection{DP procedure: Robust test construction via differentially privacy}
\label{subsec:dptest}

Differential privacy (DP) is closely related to statistical robustness \citep{dwork2009differential,avella2020}. 
This connection naturally motivates us to leverage the recent advances in differential privatisation of hypothesis tests \citep{pena2022differentially,kazan2023test,kim2023differentially} for our data corruption setting. 
Consider any valid $\varepsilon$-DP hypothesis test with adjusted lower level $\alpha e^{-r\varepsilon}$ and pure DP parameter $\varepsilon>0$. 
Under the null, the DP group privacy property \citep[Theorem 2.2]{dwork2014algorithmic} ensures that
\begin{equation}
\label{eq:valid_dp}
\begin{aligned}
	&\sup_{P_0\in\mathcal{P}_0}\mP_{P_0} \bigl(\textrm{reject } \mathcal{H}_0\,\given\, r \textrm{ corrupted data} \bigr) \\
	~\leq~ &e^{r\varepsilon} \sup_{P_0\in\mathcal{P}_0}\mP_{P_0} \bigl(\textrm{reject } \mathcal{H}_0 \,\given\, \textrm{uncorrupted data} \bigr)\\
	~\leq~ &\alpha.
\end{aligned}
\end{equation}
This guarantees that any valid $\varepsilon$-DP test with adjusted level $\alpha e^{-r\varepsilon}$ is well-calibrated under data corruption in the sense that its type I error is controlled by $\alpha$.

We choose to focus on the differential privatisation approach of \citet{kim2023differentially} for permutation tests, as it benefits from stronger theoretical guarantees.
In particular, their resulting DP tests are guaranteed to be valid at every sample size \citep[Theorem 1]{kim2023differentially}, so type I error control under data corruption of \Cref{eq:valid_dp} holds non-asymptotically. 
This is actually crucial as the corruption and privacy parameters $r$ and $\varepsilon$ usually depend on the sample size. Consequently, the adjusted level $\alpha e^{-r\varepsilon}$ also depends on the sample size, which can modify the asymptotic behaviour of type I error control.
We simply refer to this test construction based on \citet{kim2023differentially} with the adjusted level as the DP procedure (see details in \Cref{alg:dp}).

The consistency of the DP test against data corruption can be guaranteed (\emph{i.e.}, for a fixed alternative distribution, the test power converges to one asymptotically) under minimal conditions.
Such a result is presented in \Cref{sec:dp_results}, alongside with a power separation analysis of the two-sample dpMMD and independence \mbox{dpHSIC} tests, constructed from \Cref{alg:dp}.


\begin{minipage}{0.49\textwidth}
	\begin{algorithm}[H]\raggedright \caption{Robust DP procedure \\ Adapted from \citealp{kim2023differentially}}
	\label{alg:dp}
		\textbf{Inputs:} Data $\mathcal{X}_n$, robustness $r$, level $\alpha$, statistic \\ $T$, permutation number $B$, privacy $\varepsilon$.\\
	\vskip .5em
		Generate i.i.d. $\zeta_0,\dots,\zeta_B \sim \mathsf{Laplace}(0,1)$. \\
		Generate i.i.d. permutations $\bpi_1,\dots,\bpi_B$ of $[n]$. \\
		Set $\bpi_0=\mathrm{Id}$ and compute global sensitivity $\Delta_T$. \\
		Compute $M_i = T(\mathcal{X}_n^{\bpi_i}) + 2 \zeta_i\Delta_T \varepsilon^{-1} $, $i\in[B]_0$. \\
		Compute $(1\!-\!\alpha e^{-r\varepsilon})$-quantile $q$ of $M_0,\dots,M_B$. \\
	\vskip .5em
	\textbf{Output}: Reject $\mathcal{H}_0$ if $M_0 > q$.
\end{algorithm}
\end{minipage}

\section{EXPERIMENTS}
\label{sec:experiments}

\begin{figure*}[t]
\centering
\includegraphics[width=\textwidth]{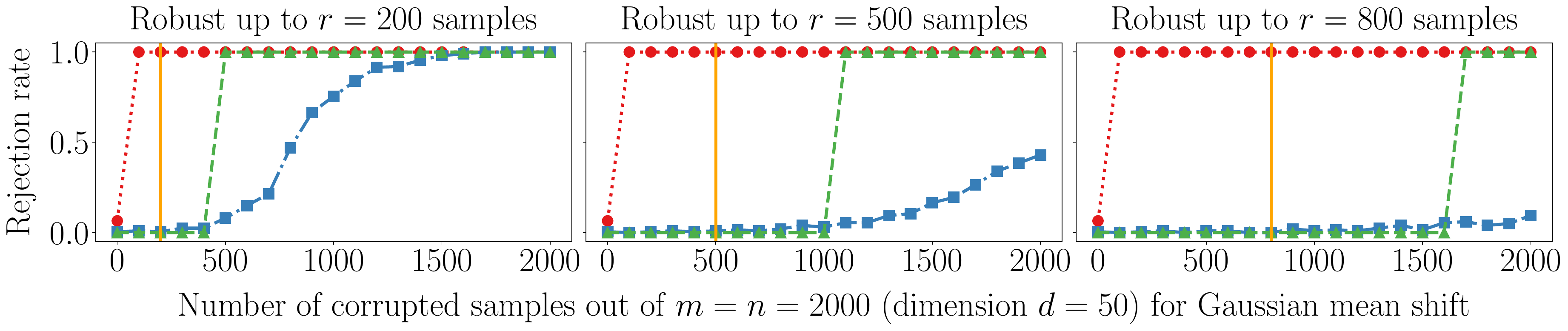}

\medskip

\includegraphics[width=\textwidth]{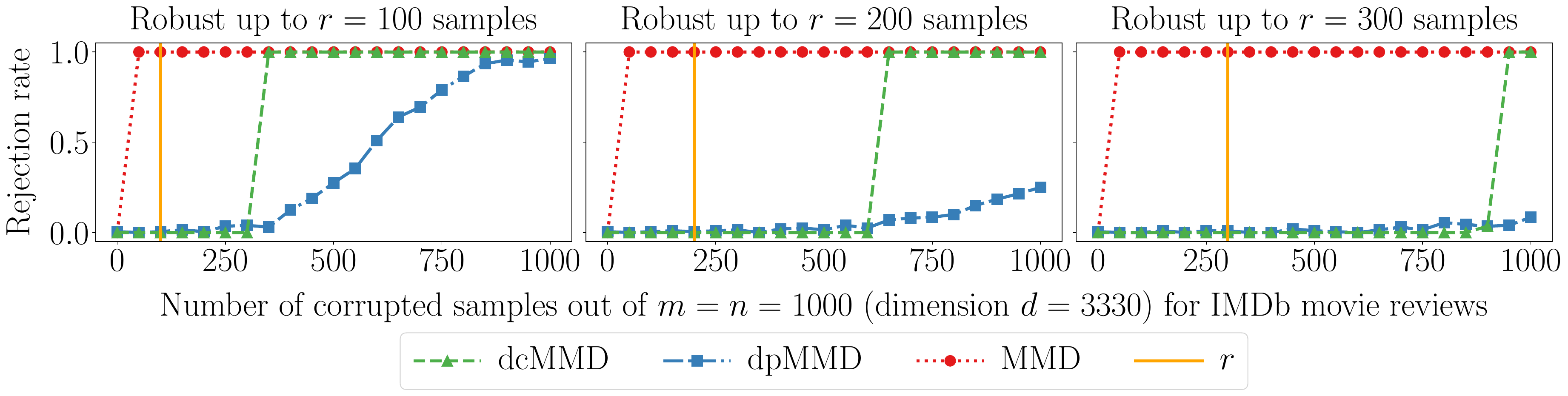}
\caption{%
Two-sample experiments robust up to $r$ corrupted samples.
To have valid level, a robust test needs to control the rejection rate by $\alpha=0.05$ when fewer than $r$ samples are corrupted. 
To be powerful, the robust test needs to have a high rejection rate when more than $r$ samples are corrupted.
\emph{(Top row: Gaussian mean shift)}
Both samples are originally i.i.d.~drawn from $\mathsf{Gaussian}(0,\nicefrac{1}{10},50)$, entries of one sample are corrupted being replaced by samples from $\mathsf{Gaussian}(1000,\nicefrac{1}{10},50)$.
\emph{(Bottom row: IMDb movie reviews)}
Both samples originally consist of movie reviews (using a bag of 3330 words representation).
Corrupted entries for one sample are replaced by samples from $\mathsf{Geometric}(0.05,3330)$.
\label{fig:mmd_experiments}
}
\end{figure*}

In this section,
we provide simulation results for two-sample and independence testing on both synthetic and real-world data, focusing on the robustness properties of the designed tests.
Our initial focus is to demonstrate that the robust tests control the type I error whenever $r$ or fewer entries have been corrupted, a task which the non-robust tests fail to accomplish. 
Assuming this control, the aim is then to have test power, \emph{e.g.}, to detect when strictly more than $r$ samples have been corrupted.

We stress that the DC and DP methods (\Cref{alg:dc,alg:dp}) are calibrated to control the type I error even under the most severe corruption scenarios.
Under the strongest $c$-corruption scenario where $c$ denotes the number of corrupted data points, we expect the rejection rate of the robust tests with robustness parameter $r$ to be exactly at $\alpha$ when $c=r$ (as the bounds in \Cref{eq:valid_dp,eq:valid_dc} are tight in this setting). We also expect test power to start increasing as soon as $c>r$. 
In practice, and in our experiments, however, we typically encounter milder corruption scenarios under which the robust tests are conservative for $c\leq r$, resulting in a type I error substantially smaller than $\alpha$. This implies a delay in the observed increase of test power, which sometimes happens only for values of $c$ much larger than $r$.

\begin{figure*}[h]
\centering
\includegraphics[width=\textwidth]{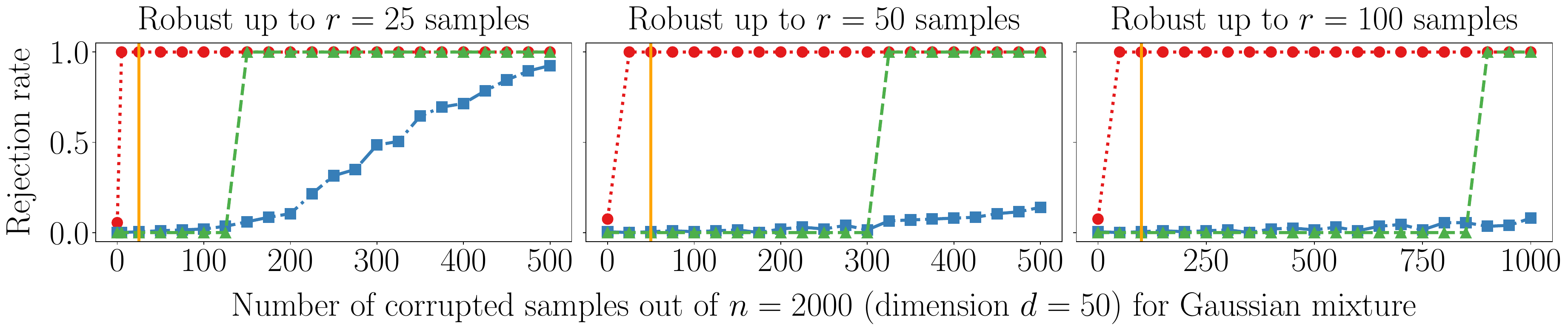}

\medskip

\includegraphics[width=\textwidth]{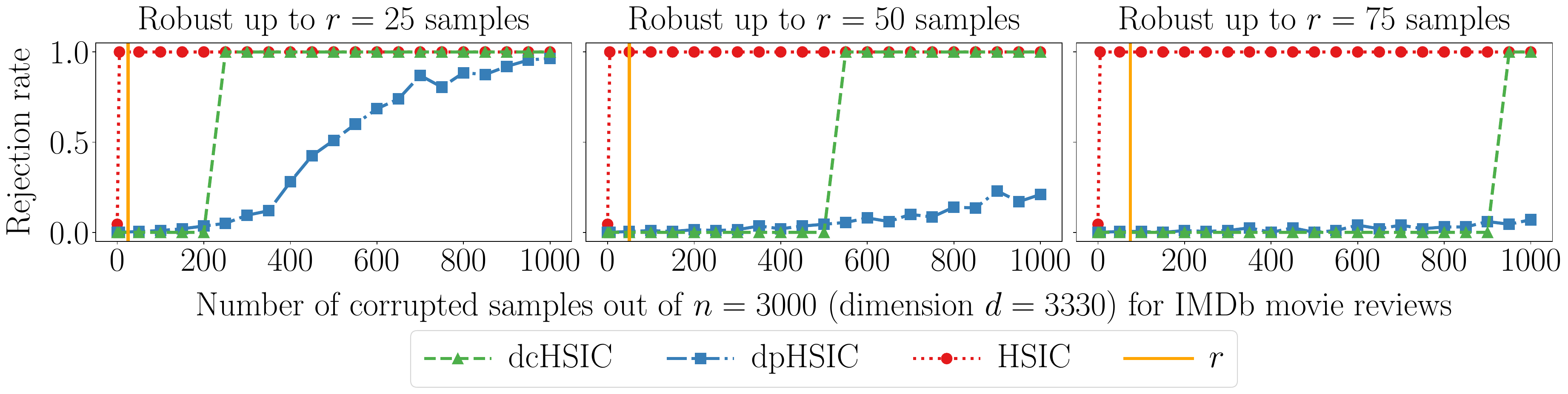}
\caption{%
Independence experiments robust up to $r$ corrupted samples.
To have valid level, a robust test needs to control the rejection rate by $\alpha=0.05$ when fewer than $r$ samples are corrupted. 
To be powerful, the robust test needs to have a high rejection rate when more than $r$ samples are corrupted.
\emph{(Top row: Gaussian mixture)} 
Paired samples $(X,Y)$ are originally i.i.d.~drawn from two $\mathsf{Gaussian}(0,\nicefrac{1}{10},50)$.
Corrupted samples are replaced by $(X, X + \epsilon)$ where $\epsilon \sim \mathsf{Gaussian}(0,\nicefrac{1}{10},50)$ and where $X\sim\mathsf{Gaussian}(s1000,\nicefrac{1}{10},50)$ with $s=1$ for half of the corrupted samples and $s=-1$ for the other half.
\emph{(Bottom row: IMDb movie reviews)}
Paired samples $(X,Y)$ originally consist of two independent reviews (represented using a bag of 3330 words).
Corrupted samples are replaced by $(X + s, X + s + \epsilon)$ where $X\sim\mathsf{Geometric}(0.05,3330)$, $\epsilon \sim \mathsf{Gaussian}(0, \nicefrac{1}{10},3330)$ and with $s=0$ for half of the corrupted samples and $s=5$ for the other half.
\label{fig:hsic_experiments}
}
\end{figure*}

In the two-sample setting, we compare the rejection rates achieved by our proposed robust dcMMD test against the dpMMD test adapted from \citet{kim2023differentially}, as well as by the non-robust permutation-based MMD test.
Similarly, we compare the performance of all three dcHSIC, dpHSIC and HSIC tests for the independence testing problem.
All tests are run using Gaussian kernels and 500 permutations.
For both frameworks, we consider synthetic Gaussian  simulations (mean shift and mixture), and experiments based on the real-world IMDb `Large Movie Review Dataset' of \citet{maas2011learning} (see captions of \Cref{fig:mmd_experiments,fig:hsic_experiments} for experimental details, rejection rates plotted are averaged over 200 repetitions).

The same trends are observed across all experiments of \Cref{fig:mmd_experiments,fig:hsic_experiments}.
First, we note that all tests control the type I error at $\alpha=0.05$ in the uncorrupted setting.
As soon as some samples are corrupted, the MMD and HSIC tests achieve rejection rate 1, \emph{i.e.}, these tests are not robust to data corruption.
In contrast, we observe that the DC and DP tests control the rejection rate to be lower than $\alpha$ when at most $r$ samples are corrupted, as theoretically guaranteed by \Cref{res:dc_consistent,apres:dp_consistent_ap}.

In the alternative regime with more than $r$ corrupted samples, the DC and DP tests behave differently, with our proposed DC tests clearly always achieving higher power than the DP tests adapted from \citet{kim2023differentially}.
The DC rejection rate rapidly increases to 1 after some threshold is reached in the number of corrupted samples, which is a desired property: the rejection rate is controlled to be lower than $\alpha$ when robustness is needed, and the power becomes 1 when robustness is no longer required.
We observe that the DP rejection rate starts increasing around the same threshold as for DC, but, unlike DC, its growth is slow and gradual.  
Overall, while both procedures control the type I error under data corruption, the DC tests are much more powerful than the DP tests.

As aforementioned, the fact that this threshold can sometimes be larger than $r$ can be explained by the design of the tests to be robust to any $r$ corruption processes, including stronger ones than those used in these experiments.
Nonetheless, we highlight the strong performance of the DC tests. For example, dcMMD is able to detect differences in distributions using only $2000$ samples while being robust to the corruption of up to $800$ of these samples (see \Cref{fig:mmd_experiments}), which is remarkable.


We note that the experiments in \Cref{fig:mmd_experiments,fig:hsic_experiments} are presented as the null hypothesis holding and being corrupted towards some alternative hypothesis (\emph{e.g.}, reading the plots from left to right).
They can equivalently be interpreted as some alternative hypothesis holding and being corrupted towards the null hypothesis (\emph{e.g.}, reading the plots from right to left).

\section{CONCLUSION}
\label{sec:conclusion}
In this paper, we introduced a general approach for constructing robust permutation tests under data corruption.
This method, called DC procedure and presented in \Cref{alg:dc}, avoids the use of random noise injection (required for the DP procedure of \citet{kim2023differentially} presented in \Cref{alg:dp}), making it more suitable for settings where reproducibility is critical. 
By employing the permutation principle, under the exchangeability assumption, both approaches ensure non-asymptotic validity under $r$ data corruption.

In terms of power guarantees, we established the minimal conditions under which the DC test maintains consistency under data corruption against any fixed alternatives. 
We illustrated our general frameworks in the context of kernel two-sample and independence testing, and showed that our kernel robust DC tests achieve minimax optimal power in terms of the MMD and HSIC metrics. 
Additionally, we demonstrated that in low corruption regimes, the robustness property can be attained without compromising the minimum separation rate in terms of power. 
Empirical results were presented to illustrate the finite-sample performance of our robust DC tests which significantly outperform the DP tests in terms of test power.

\textbf{Limitations and future work.} Our work opens up several fruitful directions for future research. As illustrated in our empirical studies, the proposed methods might be conservative in certain settings, as they are designed to safeguard testing errors against worst-case attack scenarios. It would be interesting to consider milder  and potentially more structured attack scenarios (\emph{e.g.}, Huber's $\epsilon$-contamination model), and to refine our DC procedure to mitigate this conservative nature. Furthermore, expanding our methods to other types of  test statistics and different testing problems is worth exploring. One can also attempt to develop computationally efficient robust tests by exploiting recent advancements in time-efficient kernel tests~\citep[\emph{e.g.},][]{domingo2023compress,schrab2022efficient}. Lastly, studying minimax rates for robust testing in terms of other metrics (\emph{e.g.,} Wasserstein and $L_p$ metrics) presents an interesting  avenue for future work.


\section*{Acknowledgments}
Antonin Schrab acknowledges support from the U.K. Research and Innovation (grant
EP/S021566/1). Ilmun Kim acknowledges support from the Basic Science Research Program through the National Research Foundation of Korea (2022R1A4A1033384), and the Korea government (MSIT) (RS-2023-00211073).

\bibliography{reference}

\begin{thebibliography}{}

\bibitem[Acharya et~al., 2019]{acharya2019test}
Acharya, J., Canonne, C., Freitag, C., and Tyagi, H. (2019).
\newblock {Test without trust: Optimal locally private distribution testing}.
\newblock In {\em The 22nd International Conference on Artificial Intelligence
  and Statistics}.

\bibitem[Acharya et~al., 2021a]{acharya2021differentially}
Acharya, J., Sun, Z., and Zhang, H. (2021a).
\newblock {Differentially Private Assouad, Fano, and Le Cam}.
\newblock In {\em Algorithmic Learning Theory}, pages 48--78. PMLR.

\bibitem[Acharya et~al., 2021b]{acharya2021robust}
Acharya, J., Sun, Z., and Zhang, H. (2021b).
\newblock {Robust Testing and Estimation under Manipulation Attacks}.
\newblock In {\em International Conference on Machine Learning}, pages 43--53.
  PMLR.

\bibitem[Albert et~al., 2022]{albert2019adaptive}
Albert, M., Laurent, B., Marrel, A., and Meynaoui, A. (2022).
\newblock Adaptive test of independence based on {HSIC} measures.
\newblock {\em The Annals of Statistics}, 50(2):858--879.

\bibitem[Aliakbarpour et~al., 2019]{aliakbarpour2019private}
Aliakbarpour, M., Diakonikolas, I., Kane, D., and Rubinfeld, R. (2019).
\newblock {Private Testing of Distributions via Sample Permutations}.
\newblock {\em Advances in Neural Information Processing Systems}, 32.

\bibitem[Aliakbarpour et~al., 2018]{aliakbarpour2017differentially}
Aliakbarpour, M., Diakonikolas, I., and Rubinfeld, R. (2018).
\newblock {Differentially Private Identity and Closeness Testing of Discrete
  Distributions}.
\newblock {\em In International Conference on Machine Learning}, pages
  169--178.

\bibitem[Avella-Medina, 2020]{avella2020}
Avella-Medina, M. (2020).
\newblock {The Role of Robust Statistics in Private Data Analysis}.
\newblock {\em CHANCE}, 33(4):37–42.

\bibitem[Biggs et~al., 2024]{biggs2023mmdfuse}
Biggs, F., Schrab, A., and Gretton, A. (2024).
\newblock {MMD-FUSE: Learning and Combining Kernels for Two-Sample Testing
  Without Data Splitting}.
\newblock {\em Advances in Neural Information Processing Systems}, 36.

\bibitem[Bradbury et~al., 2018]{jax2018github}
Bradbury, J., Frostig, R., Hawkins, P., Johnson, M.~J., Leary, C., Maclaurin,
  D., Necula, G., Paszke, A., Vander{P}las, J., Wanderman-{M}ilne, S., and
  Zhang, Q. (2018).
\newblock {JAX}: composable transformations of {P}ython+{N}um{P}y programs.

\bibitem[Briol et~al., 2019]{briol2019statistical}
Briol, F.-X., Barp, A., Duncan, A.~B., and Girolami, M. (2019).
\newblock {Statistical Inference for Generative Models with Maximum Mean
  Discrepancy}.
\newblock {\em arXiv preprint arXiv:1906.05944}.

\bibitem[Cai et~al., 2017]{cai2017priv}
Cai, B., Daskalakis, C., and Kamath, G. (2017).
\newblock {Priv’it: Private and sample efficient identity testing}.
\newblock In {\em International Conference on Machine Learning}, pages
  635--644. PMLR.

\bibitem[Campbell et~al., 2018]{campbell2018differentially}
Campbell, Z., Bray, A., Ritz, A., and Groce, A. (2018).
\newblock {Differentially private ANOVA testing}.
\newblock In {\em 2018 1st International Conference on Data Intelligence and
  Security (ICDIS)}, pages 281--285. IEEE.

\bibitem[Canonne et~al., 2023]{canonne2023}
Canonne, C., Hopkins, S.~B., Li, J., Liu, A., and Narayanan, S. (2023).
\newblock {The Full Landscape of Robust Mean Testing: Sharp Separations between
  Oblivious and Adaptive Contamination}.
\newblock In {\em 2023 IEEE 64th Annual Symposium on Foundations of Computer
  Science (FOCS)}. IEEE.

\bibitem[Chatterjee and Bhattacharya, 2023]{chatterjee2023boosting}
Chatterjee, A. and Bhattacharya, B.~B. (2023).
\newblock Boosting the power of kernel two-sample tests.
\newblock {\em arXiv preprint arXiv:2302.10687}.

\bibitem[Chen et~al., 2016]{chen2016general}
Chen, M., Gao, C., and Ren, Z. (2016).
\newblock {A general decision theory for Huber’s $\epsilon$-contamination
  model}.
\newblock {\em Electronic Journal of Statistics}, 10(2):3752 -- 3774.

\bibitem[Ch{\'e}rief-Abdellatif and Alquier, 2020]{cherief2020mmd}
Ch{\'e}rief-Abdellatif, B.-E. and Alquier, P. (2020).
\newblock {MMD-Bayes: Robust Bayesian estimation via maximum mean discrepancy}.
\newblock In {\em Symposium on Advances in Approximate Bayesian Inference},
  pages 1--21. PMLR.

\bibitem[Ch{\'e}rief-Abdellatif and Alquier, 2022]{cherief2022finite}
Ch{\'e}rief-Abdellatif, B.-E. and Alquier, P. (2022).
\newblock {Finite sample properties of parametric MMD estimation: robustness to
  misspecification and dependence}.
\newblock {\em Bernoulli}, 28(1):181--213.

\bibitem[Couch et~al., 2019]{couch2019differentially}
Couch, S., Kazan, Z., Shi, K., Bray, A., and Groce, A. (2019).
\newblock Differentially private nonparametric hypothesis testing.
\newblock In {\em Proceedings of the 2019 ACM SIGSAC Conference on Computer and
  Communications Security}, pages 737--751.

\bibitem[Dellaporta and Damoulas, 2023]{dellaporta2023robust}
Dellaporta, C. and Damoulas, T. (2023).
\newblock {Robust Bayesian Inference for Berkson and Classical Measurement
  Error Models}.
\newblock {\em arXiv preprint arXiv:2306.01468}.

\bibitem[Dellaporta et~al., 2022]{dellaporta2022robust}
Dellaporta, C., Knoblauch, J., Damoulas, T., and Briol, F.-X. (2022).
\newblock {Robust Bayesian inference for simulator-based models via the MMD
  posterior bootstrap}.
\newblock In {\em International Conference on Artificial Intelligence and
  Statistics}, pages 943--970. PMLR.

\bibitem[Diakonikolas and Kane, 2019]{diakonikolas2019recent}
Diakonikolas, I. and Kane, D.~M. (2019).
\newblock {Recent Advances in Algorithmic High-Dimensional Robust Statistics}.
\newblock {\em arXiv preprint arXiv:1911.05911}.

\bibitem[Diakonikolas and Kane, 2021]{diakonikolas2021sample}
Diakonikolas, I. and Kane, D.~M. (2021).
\newblock The sample complexity of robust covariance testing.
\newblock In {\em Conference on Learning Theory}, pages 1511--1521. PMLR.

\bibitem[Diakonikolas et~al., 2017]{diakonikolas2017}
Diakonikolas, I., Kane, D.~M., and Stewart, A. (2017).
\newblock {Statistical Query Lower Bounds for Robust Estimation of
  High-Dimensional Gaussians and Gaussian Mixtures}.
\newblock In {\em 2017 IEEE 58th Annual Symposium on Foundations of Computer
  Science (FOCS)}. IEEE.

\bibitem[Domingo-Enrich et~al., 2023]{domingo2023compress}
Domingo-Enrich, C., Dwivedi, R., and Mackey, L. (2023).
\newblock {Compress then test: Powerful kernel testing in near-linear time}.
\newblock In {\em International Conference on Artificial Intelligence and
  Statistics}.

\bibitem[Dwork and Lei, 2009]{dwork2009differential}
Dwork, C. and Lei, J. (2009).
\newblock Differential privacy and robust statistics.
\newblock In {\em Proceedings of the forty-first annual ACM symposium on Theory
  of computing}, pages 371--380.

\bibitem[Dwork et~al., 2006]{dwork2006calibrating}
Dwork, C., McSherry, F., Nissim, K., and Smith, A. (2006).
\newblock Calibrating noise to sensitivity in private data analysis.
\newblock In {\em Theory of cryptography conference}, pages 265--284. Springer.

\bibitem[Dwork et~al., 2014]{dwork2014algorithmic}
Dwork, C., Roth, A., et~al. (2014).
\newblock {The Algorithmic Foundations of Differential Privacy}.
\newblock {\em Foundations and Trends{\textregistered} in Theoretical Computer
  Science}, 9(3--4):211--407.

\bibitem[Fazeli-Asl et~al., 2024]{fazeli2024semi}
Fazeli-Asl, F., Zhang, M.~M., and Lin, L. (2024).
\newblock A semi-bayesian nonparametric estimator of the maximum mean
  discrepancy measure: Applications in goodness-of-fit testing and generative
  adversarial networks.
\newblock {\em Transactions on Machine Learning Research}.

\bibitem[Fienberg et~al., 2011]{fienberg2011privacy}
Fienberg, S.~E., Slavkovic, A., and Uhler, C. (2011).
\newblock {Privacy preserving GWAS data sharing}.
\newblock In {\em 2011 IEEE 11th International Conference on Data Mining
  Workshops}, pages 628--635. IEEE.

\bibitem[Fukumizu et~al., 2008]{fukumizu2008kernel}
Fukumizu, K., Gretton, A., Sun, X., and Sch{\"o}lkopf, B. (2008).
\newblock Kernel measures of conditional dependence.
\newblock In {\em Advances in Neural Information Processing Systems}, volume~1,
  pages 489--496.

\bibitem[Gao et~al., 2018]{gao2018robust}
Gao, R., Xie, L., Xie, Y., and Xu, H. (2018).
\newblock {Robust Hypothesis Testing Using Wasserstein Uncertainty Sets}.
\newblock In {\em Advances in Neural Information Processing Systems},
  volume~31.

\bibitem[George and Canonne, 2022]{george2022robust}
George, A.~J. and Canonne, C.~L. (2022).
\newblock Robust testing in high-dimensional sparse models.
\newblock {\em Advances in Neural Information Processing Systems},
  35:16469--16480.

\bibitem[Gretton, 2015]{gretton2015simpler}
Gretton, A. (2015).
\newblock A simpler condition for consistency of a kernel independence test.
\newblock {\em arXiv preprint arXiv:1501.06103}.

\bibitem[Gretton et~al., 2012a]{gretton2012kernel}
Gretton, A., Borgwardt, K.~M., Rasch, M.~J., Sch{\"o}lkopf, B., and Smola, A.
  (2012a).
\newblock A kernel two-sample test.
\newblock {\em The Journal of Machine Learning Research}, 13(1):723--773.

\bibitem[Gretton et~al., 2005a]{gretton2005measuring}
Gretton, A., Bousquet, O., Smola, A., and Sch{\"o}lkopf, B. (2005a).
\newblock {Measuring statistical dependence with Hilbert-Schmidt norms}.
\newblock In {\em International Conference on Algorithmic Learning Theory},
  pages 63--77. Springer.

\bibitem[Gretton et~al., 2005b]{gretton2005kernel}
Gretton, A., Herbrich, R., Smola, A., Bousquet, O., and Sch{\"o}lkopf, B.
  (2005b).
\newblock Kernel methods for measuring independence.
\newblock {\em Journal of Machine Learning Research}, 6:2075--2129.

\bibitem[Gretton et~al., 2012b]{gretton2012optimal}
Gretton, A., Sejdinovic, D., Strathmann, H., Balakrishnan, S., Pontil, M.,
  Fukumizu, K., and Sriperumbudur, B.~K. (2012b).
\newblock Optimal kernel choice for large-scale two-sample tests.
\newblock {\em Advances in Neural Information Processing Systems}, 25.

\bibitem[Hagrass et~al., 2024]{hagrass2022spectral}
Hagrass, O., Sriperumbudur, B.~K., and Li, B. (2024).
\newblock {Spectral Regularized Kernel Two-Sample Tests}.
\newblock {\em The Annals of Statistics}, 52(3):1076--1101.

\bibitem[Hemerik and Goeman, 2018]{hemerik2018exact}
Hemerik, J. and Goeman, J. (2018).
\newblock Exact testing with random permutations.
\newblock {\em Test}, 27(4):811--825.

\bibitem[Huber, 1964]{huber1964}
Huber, P.~J. (1964).
\newblock {Robust Estimation of a Location Parameter}.
\newblock {\em The Annals of Mathematical Statistics}, 35(1):73–101.

\bibitem[Kazan et~al., 2023]{kazan2023test}
Kazan, Z., Shi, K., Groce, A., and Bray, A.~P. (2023).
\newblock {The test of tests: A framework for differentially private hypothesis
  testing}.
\newblock In {\em International Conference on Machine Learning}, pages
  16131--16151. PMLR.

\bibitem[Kim, 2021]{kim2021comparing}
Kim, I. (2021).
\newblock Comparing a large number of multivariate distributions.
\newblock {\em Bernoulli}, 27(1):419--441.

\bibitem[Kim and Schrab, 2023]{kim2023differentially}
Kim, I. and Schrab, A. (2023).
\newblock {Differentially Private Permutation Tests: Applications to Kernel
  Methods}.
\newblock {\em arXiv preprint arXiv:2310.19043}.

\bibitem[Le~Cam, 1973]{lecam1973convergence}
Le~Cam, L. (1973).
\newblock {Convergence of Estimates Under Dimensionality Restrictions}.
\newblock {\em The Annals of Statistics}, 1(1):38--53.

\bibitem[Le~Cam, 2012]{le2012asymptotic}
Le~Cam, L. (2012).
\newblock {\em {Asymptotic Methods in Statistical Decision Theory}}.
\newblock Springer Science \& Business Media.

\bibitem[Legramanti et~al., 2025]{legramanti2022concentration}
Legramanti, S., Durante, D., and Alquier, P. (2025).
\newblock {Concentration of discrepancy-based approximate Bayesian computation
  via Rademacher complexity}.
\newblock {\em The Annals of Statistics}, 53(1):37--60.

\bibitem[Lehmann and Romano, 2005]{lehmann2005testing}
Lehmann, E.~L. and Romano, J.~P. (2005).
\newblock {\em {Testing Statistical Hypotheses}}, volume~3.
\newblock Springer.

\bibitem[Li et~al., 2023]{li2023robustness}
Li, M., Berrett, T.~B., and Yu, Y. (2023).
\newblock On robustness and local differential privacy.
\newblock {\em The Annals of Statistics}, 51(2):717--737.

\bibitem[Li and Yuan, 2024]{li2019optimality}
Li, T. and Yuan, M. (2024).
\newblock {On the Optimality of Gaussian Kernel Based Nonparametric Tests
  against Smooth Alternatives}.
\newblock {\em Journal of Machine Learning Research}, 25(334):1--62.

\bibitem[Maas et~al., 2011]{maas2011learning}
Maas, A.~L., Daly, R.~E., Pham, P.~T., Huang, D., Ng, A.~Y., and Potts, C.
  (2011).
\newblock Learning word vectors for sentiment analysis.
\newblock In {\em Proceedings of the 49th Annual Meeting of the Association for
  Computational Linguistics: Human Language Technologies}, pages 142--150,
  Portland, Oregon, USA.

\bibitem[Pe{\~n}a and Barrientos, 2022]{pena2022differentially}
Pe{\~n}a, V. and Barrientos, A.~F. (2022).
\newblock {Differentially Private Hypothesis Testing with the Subsampled and
  Aggregated Randomized Response Mechanism}.
\newblock {\em arXiv preprint arXiv:2208.06803}.

\bibitem[Raj et~al., 2020]{raj2020differentially}
Raj, A., Law, H. C.~L., Sejdinovic, D., and Park, M. (2020).
\newblock A differentially private kernel two-sample test.
\newblock {\em Lecture Notes in Computer Science}, 11906.

\bibitem[Rogers and Kifer, 2017]{rogers2017new}
Rogers, R. and Kifer, D. (2017).
\newblock A new class of private chi-square hypothesis tests.
\newblock In {\em Artificial Intelligence and Statistics}, pages 991--1000.
  PMLR.

\bibitem[Romano and Wolf, 2005]{romano2005exact}
Romano, J.~P. and Wolf, M. (2005).
\newblock Exact and approximate stepdown methods for multiple hypothesis
  testing.
\newblock {\em Journal of the American Statistical Association},
  100(469):94--108.

\bibitem[Schrab et~al., 2022a]{schrab2022ksd}
Schrab, A., Guedj, B., and Gretton, A. (2022a).
\newblock {KSD Aggregated Goodness-of-fit Test}.
\newblock {\em Advances in Neural Information Processing Systems},
  35:32624--32638.

\bibitem[Schrab et~al., 2023]{schrab2021mmd}
Schrab, A., Kim, I., Albert, M., Laurent, B., Guedj, B., and Gretton, A.
  (2023).
\newblock {MMD Aggregated Two-Sample Test}.
\newblock {\em Journal of Machine Learning Research}, 24(194):1--81.

\bibitem[Schrab et~al., 2022b]{schrab2022efficient}
Schrab, A., Kim, I., Guedj, B., and Gretton, A. (2022b).
\newblock {Efficient Aggregated Kernel Tests using Incomplete
  {$U$}-statistics}.
\newblock {\em Advances in Neural Information Processing Systems},
  35:18793--18807.

\bibitem[Shekhar et~al., 2022]{shekhar2022two}
Shekhar, S., Kim, I., and Ramdas, A. (2022).
\newblock A permutation-free kernel two-sample test.
\newblock {\em Advances in Neural Information Processing Systems}.

\bibitem[Shekhar et~al., 2023]{shekhar2022ind}
Shekhar, S., Kim, I., and Ramdas, A. (2023).
\newblock A permutation-free kernel independence test.
\newblock {\em Journal of Machine Learning Research}, 24(369):1--68.

\bibitem[Sun and Zou, 2021]{sun2021data}
Sun, Z. and Zou, S. (2021).
\newblock {A Data-Driven Approach to Robust Hypothesis Testing Using Kernel MMD
  Uncertainty Sets}.
\newblock In {\em 2021 IEEE International Symposium on Information Theory
  (ISIT)}. IEEE.

\bibitem[Sun and Zou, 2022]{sun2022robust}
Sun, Z. and Zou, S. (2022).
\newblock Robust hypothesis testing with kernel uncertainty sets.
\newblock In {\em 2022 IEEE International Symposium on Information Theory
  (ISIT)}, pages 3309--3314. IEEE.

\bibitem[Sun and Zou, 2023]{sun2023kernel}
Sun, Z. and Zou, S. (2023).
\newblock Kernel robust hypothesis testing.
\newblock {\em IEEE Transactions on Information Theory}, 69(10):6619--6638.

\bibitem[Wang and Xie, 2022]{wang2022data}
Wang, J. and Xie, Y. (2022).
\newblock {A Data-Driven Approach to Robust Hypothesis Testing Using Sinkhorn
  Uncertainty Sets}.
\newblock In {\em 2022 IEEE International Symposium on Information Theory
  (ISIT)}. IEEE.

\bibitem[Yamada et~al., 2019]{yamada2019post}
Yamada, M., Wu, D., Tsai, Y.-H.~H., Takeuchi, I., Salakhutdinov, R., and
  Fukumizu, K. (2019).
\newblock Post selection inference with incomplete maximum mean discrepancy
  estimator.
\newblock {\em International Conference on Learning Representations}.

\bibitem[Zaremba et~al., 2013]{zaremba2013b}
Zaremba, W., Gretton, A., and Blaschko, M. (2013).
\newblock {B-test: A non-parametric, low variance kernel two-sample test}.
\newblock {\em Advances in Neural Information Processing Systems}, 26.

\bibitem[Zhao and Meng, 2015]{zhao2015fastmmd}
Zhao, J. and Meng, D. (2015).
\newblock {FastMMD: Ensemble of circular discrepancy for efficient two-sample
  test}.
\newblock {\em Neural Computation}, 27(6):1345--1372.

\end{thebibliography}

\section*{Checklist}



\begin{enumerate}

\item For all models and algorithms presented, check if you include:
\begin{enumerate}
  \item A clear description of the mathematical setting, assumptions, algorithm, and/or model. Yes
  \item An analysis of the properties and complexity (time, space, sample size) of any algorithm. Yes
  \item (Optional) Anonymized source code, with specification of all dependencies, including external libraries. Yes
\end{enumerate}

\item For any theoretical claim, check if you include:
\begin{enumerate}
  \item Statements of the full set of assumptions of all theoretical results. Yes
  \item Complete proofs of all theoretical results. Yes
  \item Clear explanations of any assumptions. Yes
\end{enumerate}

\item For all figures and tables that present empirical results, check if you include:
\begin{enumerate}
  \item The code, data, and instructions needed to reproduce the main experimental results (either in the supplemental material or as a URL). Yes
  \item All the training details (e.g., data splits, hyperparameters, how they were chosen). Yes
        \item A clear definition of the specific measure or statistics and error bars (e.g., with respect to the random seed after running experiments multiple times). Not Applicable (outputs are binary so error bars are deterministic given the number of repetitions)
        \item A description of the computing infrastructure used. (e.g., type of GPUs, internal cluster, or cloud provider). Yes
\end{enumerate}

\item If you are using existing assets (e.g., code, data, models) or curating/releasing new assets, check if you include:
\begin{enumerate}
  \item Citations of the creator If your work uses existing assets. Yes
  \item The license information of the assets, if applicable. Yes
  \item New assets either in the supplemental material or as a URL, if applicable. Yes
  \item Information about consent from data providers/curators. Yes
  \item Discussion of sensible content if applicable, e.g., personally identifiable information or offensive content. Not Applicable
\end{enumerate}

\item If you used crowdsourcing or conducted research with human subjects, check if you include:
\begin{enumerate}
  \item The full text of instructions given to participants and screenshots. Not Applicable
  \item Descriptions of potential participant risks, with links to Institutional Review Board (IRB) approvals if applicable. Not Applicable
  \item The estimated hourly wage paid to participants and the total amount spent on participant compensation. Not Applicable
\end{enumerate}

\end{enumerate}

\newpage
\appendix
\onecolumn 
\pagestyle{empty}
\aistatstitle{Supplementary Materials \\ Robust Kernel Hypothesis Testing under Data Corruption}

\vspace{-1cm}

In this supplementary material, we
discuss computational resources (\Cref{sec:resources}), 
present assumptions (\Cref{sec:assumptions}), derive proofs of the theoretical results presented in the main text (\Cref{sec:proofs}), 
and offer a power analysis of the DP tests (\Cref{sec:dp_results}).

\section{EXPERIMENTAL RESOURCES}
\label{sec:resources}

Reproducible code in JAX \citep{jax2018github} is available at \url{https://github.com/antoninschrab/dckernel-paper} under the MIT License. Experiments were run on a 24Gb RTX A5000 GPU, total compute time is of the order of ten hours. For the DP tests, we adapt the \href{https://github.com/antoninschrab/dpkernel-paper/}{implementation} of \citet{kim2023differentially} published under the MIT License. We use the publicly available \href{https://ai.stanford.edu/~amaas/data/sentiment/}{IMDb dataset} of \citet{maas2011learning}.

\section{ASSUMPTIONS}
\label{sec:assumptions}
This section collects several technical assumptions used in the main text to streamline our presentation. We first make the following assumptions on kernels. 
\begin{assumption}[Conditions on kernel for \Cref{res:mmd_separation}] \label{assumption: two-sample kernel} Let $k: \mathbb{S} \times \mathbb{S} \mapsto \mathbb{R}$ be a reproducing kernel defined on $\mathbb{S}$. 
	\begin{enumerate}[leftmargin=*,nolistsep]
		\item[(i)] Assume that the kernel $k$ is characteristic, non-negative, and bounded everywhere by a constant $K$, i.e., $0 \leq k(x,y) \leq K$ for all $x,y \in \mathbb{S}$. \\[-0.5em]
		\item[(ii)] Setting $\mathbb{S} = \mathbb{R}^d$, assume that the kernel $k$ is translation invariant on $\mathbb{S}$, i.e., there exists a symmetric positive definite function $\kappa$ such that $k(x,y) = \kappa(x-y)$ for all $x,y \in \mathbb{R}^d$. Moreover, assume that the kernel is non-constant in the sense that there exists a positive constant $\eta$ such that $\kappa(0) - \kappa(z) \geq \eta$ for some $z \in \mathbb{R}^d$.  
	\end{enumerate}
\end{assumption}

\smallskip 
\begin{assumption}[Conditions on kernels for \Cref{res:hsic_separation}] \label{assumption: independence kernel} Let $k: \mathbb{S}_Y \times \mathbb{S}_Y \mapsto \mathbb{R}$ and $\ell: \mathbb{S}_Z \times \mathbb{S}_Z \mapsto \mathbb{R}$ be reproducing kernels defined on $\mathbb{S}_Y\!$ and $\mathbb{S}_Z$, respectively. 
	\begin{enumerate}[leftmargin=*,nolistsep]
		\item[(i)] Assume that the kernels $k$ and $\ell$ are characteristic, non-negative, and bounded everywhere by constants $K$ and $L$, respectively, i.e., $0 \leq k(y,y') \leq K$ for all $y,y' \in \mathbb{S}_Y$ and $0 \leq \ell(z,z') \leq L$ for all $z,z' \in \mathbb{S}_Z$. \\[-0.5em]
		\item[(ii)] Setting $\mathbb{S}_Y = \mathbb{R}^{d_Y}$ and $\mathbb{S}_Z = \mathbb{R}^{d_Z}$, assume that the kernels $k$ and $\ell$ are translation invariant on $\mathbb{S}_Y$ and $\mathbb{S}_Z$, i.e., there exist symmetric positive definite functions $\kappa_Y$ and $\kappa_Z$ such that $k(y,y') = \kappa_Y(y-y')$ for all $y,y' \in \mathbb{R}^{d_Y}$ and $\ell(z,z') = \kappa_Z(z-z')$ for all $z,z' \in \mathbb{R}^{d_Z}$. Moreover, assume that the kernels are non-constant in the sense that there exist positive constants $\eta_Y,\eta_Z$ such that $\kappa_Y(0) - \kappa_Y(y_0) \geq \eta_Y$ for some $y_0 \in \mathbb{R}^{d_Y}$ and $\kappa_Z(0) - \kappa_Z(z_0) \geq \eta_Z$ for some $z_0 \in \mathbb{R}^{d_Z}$.  
	\end{enumerate}
\end{assumption}
	
As discussed in \citet{kim2023differentially}, commonly used kernels, such as the Gaussian kernel and the Laplace kernel, satisfy the above assumptions. We next describe the assumption for $B$, the number of permutations, used in \Cref{res:mmd_separation,apres:mmd_separation_ap,res:hsic_separation,apres:hsic_separation_ap}.

\begin{assumption} \label{assumption: mmd_separation}
	We make the following assumptions on the number of permutations $B$ for \Cref{res:mmd_separation,apres:mmd_separation_ap,res:hsic_separation,apres:hsic_separation_ap}:
	\begin{itemize}[leftmargin=*,nolistsep]
		\item Permutation numbers for dpMMD/dpHSIC: $B \geq 6\alpha_{\mathtt{dp}}^{-1} \log(2/\beta_{\mathtt{dp}})$ where $\alpha_{\mathtt{dp}} = e^{-r\varepsilon} \alpha$ and $\beta_{\mathtt{dp}} = e^{-r\varepsilon} \beta$.
		\item Permutation numbers for dcMMD/dcHSIC: $B \geq 3 \alpha^{-2}\bigl\{ \log\bigl(8/\beta\bigr) + \alpha (1-\alpha) \}$.
	\end{itemize}
\end{assumption}

\clearpage

The reason for the difference in the required number of permutations between the DP test and the DC test stems from their reliance on different techniques. Specifically, we can guarantee that $M_0,M_1,\ldots,M_B$ in \Cref{alg:dp} are all distinct with probability one, due to the injection of continuous noise to the test statistics. This property allows us to employ the multiplicative Chernoff inequality as in \citet[Lemma 21]{kim2023differentially}. On the other hand, there is no guarantee that $T_0,T_1,\ldots,T_B$ are distinct in \Cref{alg:dc}, for which we apply the Dvoretzky--Kiefer--Wolfowitz inequality to analyse the random permutation distribution, as in \citet[Proposition 4]{schrab2021mmd}. If we randomly break ties in $T_0,T_1,\ldots,T_B$, the condition for dcMMD/dcHSIC can be improved to $6\alpha^{-1} \log(2/\beta)$.


\section{PROOFS OF DC PROCEDURE MINIMAX OPTIMALITY}
\label{sec:proofs}
This section collects the proofs of the technical results in the main text. 

\paragraph{Additional notation.} Let $X \sim P$ and $Y \sim Q$. We use $d_{\mathrm{TV}}(P,Q)$ to denote the total variation (TV) distance between $P$ and $Q$. With an abuse of notation, we also use $d_{\mathrm{TV}}(X,Y)$ to denote $d_{\mathrm{TV}}(P,Q)$. As in the main text, the subscript $P$ in $\mP_P$ is used to emphasise that the underlying data are generated from the distribution $P$. We often omit the dependence on $P$ in $\mP_P$ whenever it is implicitly clear from the context. We stress that the probabilities are also taken with respect to $\bpi$, that is, with respect to the uniformly random draw of permutations.

\subsection{Proof of \Cref{res:dc_consistent}} 

\dcconsistent*

\begin{proof}
(i) To simplify the notation, define $U_0 = T(\tXn)$, $U_i = T(\tXn^{\bpi_i})$ for $i \in [B]$, (\emph{i.e.}, the test statistics based on the uncorrupted dataset $\tXn$ and its permuted counterparts $\tXn^{\bpi_i}$) and denote the $(1-\alpha)$-quantile of $U_0,U_1,\ldots,U_B$ as $\widetilde{q}$. Equipped with this notation, observe that $|U_i - T_i| \leq r \Delta_T$ for each $i \in [B]_0$, which follows by the repeated use of the triangle inequality and by using the definition of global sensitivity, \emph{i.e.}, $\Delta_T = \sup_{\bpi \in \boldsymbol{\Pi}_n} \sup_{\substack{\mathcal{X}_{n},{\mathcal{Y}}_{n}\,:\,d_{\mathrm{ham}}(\mathcal{X}_{n},{\mathcal{Y}}_{n}) \leq 1}} | T(\mathcal{X}_{n}^{\bpi}) - T({\mathcal{Y}}_{n}^{\bpi})|$. This inequality yields
\begin{align}
\label{eq:valid_dc}
	\mathds{1}(T_0 > q + 2r\Delta_T) \leq \mathds{1}(U_0 + r\Delta_T >  \widetilde{q} - r\Delta_T + 2r \Delta_T ) = \mathds{1}(U_0 > \tilde{q}). 
\end{align}
Since $U_0,U_1,\ldots,U_B$ are exchangeable under the null, we have $\mP_{P_0}(T_0 > q + 2r\Delta_T) \leq \mP_{P}(U_0 > \tilde{q}) \leq \alpha$ for any $P_0 \in \mathcal{P}_0$ \citep[Lemma 1]{romano2005exact}, which proves the validity result. 

\medskip 
	
(ii) Using the same notation, we first note that the type II error of the DC test is bounded above as follows
\begin{align} \label{Eq: type II error conversion}
    \mP_{P_1}(T_0 \leq q + 2r\Delta_T) \leq \mP_{P_1}(U_0 - r \Delta_T \leq \widetilde{q} + r \Delta_T + 2r\Delta_T ) = \mP_{P_1}(U_0 \leq \widetilde{q} + 4r \Delta_T),
\end{align}
where we use the fact that $|U_i - T_i| \leq r \Delta_T$ for each $i \in [B]_0$. Therefore, in order to prove that the DC test is consistent, it suffices to show that the above upper bound for the type II error converges to zero as $n \rightarrow \infty$ under the given conditions. Now, denoting $\overline{U}_i = U_i + 4r\Delta_T$ for $i \in [B]_0$, 
by definitions of quantiles, we obtain
\begin{align*}
    \mathds{1} \biggl(\frac{1}{B+1} \biggl\{\sum_{i=1}^{B} \mathds{1}(U_0 \leq \overline{U}_i) + 1\biggr\} \leq \alpha\biggr)  & = \mathds{1} \biggl(\frac{1}{B+1} \sum_{i=0}^{B} \mathds{1}(U_0 \leq \overline{U}_i) \leq \alpha\biggr) \\
    & = \mathds{1}(U_0 > \widetilde{q} + 4r \Delta_T).
\end{align*}
Using this alternative representation of the test in conjunction with \citet[][Lemma 8]{kim2023differentially}, it follows that
\begin{align*}
    \lim_{n \rightarrow \infty} \mP_{P_1}(U_0 > \widetilde{q} + 4r \Delta_T) = 1 \quad \text{equivalently} \quad \lim_{n \rightarrow \infty} \mP_{P_1}(U_0 \leq \widetilde{q} + 4r \Delta_T) = 0,
\end{align*}
if $\lim_{n\rightarrow \infty} \mP_{P_1}(U_0 \leq \overline{U}_1)  = \lim_{n\rightarrow \infty} \mP_{P_1}(T(\tXn) \leq T(\tXn^{\bpi}) + 4r \Delta_T) = 0$ and $\min_{n \in \mathbb{N}} B_n > \alpha^{-1} - 1$. This completes the proof of \Cref{res:dc_consistent}.
\end{proof}

\subsection{Proof of \Cref{res:mmd_consistent}}

\mmdconsistent*

\begin{proof}
In order to prove the consistency of dcMMD, it is enough to verify the condition in \Cref{res:dc_consistent}:
$$
\lim_{n \rightarrow \infty} \mP_{P_1}\biggl(\,\widehat{\mathrm{MMD}}(\widetilde{\mathcal{X}}_{n+m})  ~>~ \widehat{\mathrm{MMD}}(\widetilde{\mathcal{X}}_{n+m}^{\bpi})+ 4r\frac{\sqrt{2K}}{n} \biggr) = 1.
$$
As $r/n \to 0$, we need to verify that 
$$
\lim_{n \rightarrow \infty} \mP_{P,Q}\Bigl(\,\widehat{\mathrm{MMD}}(\widetilde{\mathcal{X}}_{n+m}) >~ \widehat{\mathrm{MMD}}(\widetilde{\mathcal{X}}_{n+m}^{\bpi}) \Bigr) = 1,
$$ 
which holds by the result of \citet[][Theorem 5]{kim2023differentially}. 
This proves that dcMMD is consistent.
\end{proof}

\subsection{Proof of \Cref{res:mmd_separation}}

\mmdseparation*

\begin{proof}
We emphasise that from \Cref{res:dc_consistent}, dcMMD controls the non-asymptotic type I error rate at level $\alpha$. Hence, we focus on the type II error guarantees.

(i)
As shown in \Cref{Eq: type II error conversion}, the type II error of the DC test based on corrupted data $\Xnm$ is bounded above by the type II error of the modified DC test based on uncorrupted data $\tXnm$. This modified DC test uses the cutoff value $\widetilde q + 4r\Delta_T$ instead of $q + 2r\Delta_T$. This slightly inflated cutoff value only affects the constant factor in the minimum uniform separation. We therefore assume that the data are \emph{not corrupted} throughout the proof and derive the minimum uniform separation for the dcMMD test.

Denoting the $(1-\alpha/2)$-quantile of the full permutation distribution as
\begin{align*}
    q_{\infty,1-\alpha/2} = \inf\bigg\{u \in \mathbb{R} : 1 - \frac{\alpha}{2}  \leq \frac{1}{(n+m)!} \sum_{\bpi \in \boldsymbol{\Pi}_{n+m}} \mathds{1}\bigl( \widehat{\mathrm{MMD}}(\widetilde{\mathcal{X}}_{n+m}^{\bpi})  \leq u \bigr) \bigg\},
\end{align*}
define the event that $E_1 := \{\widetilde{q} \leq q_{\infty,1-\alpha/2}\}$. With $B \geq 3 \alpha^{-2}\bigl\{ \log\bigl(8/\beta\bigr) + \alpha (1-\alpha) \}$, following the proof of \citet[][Proposition 4]{schrab2021mmd} shows that $\mP_{P,Q}(E_1) \geq 1 - \beta/2$. Moreover, \citet[][Theorem 5.1]{kim2021comparing} with the condition $n \asymp m$ ensures that 
\begin{align*}
    q_{\infty,1-\alpha/2} \leq C_{1,K} \sqrt{\frac{\log(2/\alpha)}{n}},
\end{align*}
where $C_{1,K}, C_{2,K},\ldots$ are constants only depending on $K$. Define another event 
\begin{align*}
    E_2 := \biggl\{ \bigg| \mathrm{MMD}_k(P,Q) - \widehat{\mathrm{MMD}}(\widetilde{\mathcal{X}}_{n+m}) \bigg|  \leq C_{2,K} \sqrt{\frac{\log(4/\beta)}{n}} \biggr\},
\end{align*}
which satisfies $\mP_{P,Q}(E_2) \geq 1 - \beta/2$ by \citet[][Theorem 7]{gretton2012kernel}. With the two events $E_1$ and $E_2$, holding with high probability, the type II error of the dcMMD test is bounded above as
\begin{align*}
   & \mP_{P,Q} \bigl(\textrm{dcMMD fails to reject } \mathcal{H}_0 \,\given\, \textrm{uncorrupted data} \bigr)  \\
   =~ &\mP_{P,Q}\biggl(\widehat{\mathrm{MMD}}(\widetilde{\mathcal{X}}_{n+m}) \leq \widetilde q + \frac{4r\sqrt{2K}}{n} \biggr) \\
     \leq ~ & \mP_{P,Q}\biggl( \widehat{\mathrm{MMD}}(\widetilde{\mathcal{X}}_{n+m}) \leq  C_{1,K} \sqrt{\frac{\log(2/\alpha)}{n}} + \frac{4r\sqrt{2K}}{n} \biggr) + \mP_{P,Q}(E_1^c) \\
   	\leq ~ & \mP_{P,Q}\biggl( \mathrm{MMD}_k(P,Q)- C_{2,K} \sqrt{\frac{\log(4/\beta)}{n}} \leq C_{1,K} \sqrt{\frac{\log(2/\alpha)}{n}} + \frac{4r\sqrt{2K}}{n} \biggr) \\
    & \hskip 5em + \mP_{P,Q}(E_1^c) + \mP_{P,Q}(E_2^c) \\
     \leq ~ & \mP_{P,Q}\biggl( \mathrm{MMD}_k(P,Q) \leq C_{3,K} \max \biggl\{ \sqrt{\frac{\max\{\log(e/\alpha),\log(e/\beta)\}}{n}}, \, \frac{r}{n} \biggr\}  \biggr) + \beta = \beta,
\end{align*}
where the last equality holds by taking $C_{K} > C_{3,K}$ in the theorem statement.

(ii)
We aim to prove that if the separation parameter $\rho$ in \Cref{def: MMD alternative} is smaller than the following threshold:
\begin{align*}
	\rho \leq C_{\eta} \max\biggl\{\min\biggl(\sqrt{\frac{\log(e/(\alpha+\beta))}{n}}, 1\biggr), \, \frac{r}{n} \biggr\},
\end{align*}
no test can have power greater than $1 - \beta$ uniformly over $\mathcal{P}_{\mathrm{MMD}_k}(\rho)$ where $C_\eta > 0$ is a small constant depending on $\eta$ in \Cref{assumption: two-sample kernel}. The first part of the lower bound, involving $n^{-1/2}$, was obtained in \citet[][Appendix E.10.1]{kim2023differentially}, except that we have $\log(e/(\alpha+\beta))$ instead of $\log(1/(\alpha+\beta))$. Under the condition $\alpha + \beta < 0.4$, it can be seen that both are the same rate by adjusting the constant factor $C_\eta$. Hence, it is enough to prove that the second part of the threshold, involving $r/n$.

As remarked in \citet[][Appendix E.10.1]{kim2023differentially}, the minimax separation for two-sample testing is not smaller than that for one-sample testing. Therefore, it suffices to derive the lower bound result for one-sample testing. Specifically, given i.i.d.~observations $Y_1,\ldots,Y_n$ drawn from $P$, we are interested in distinguishing $\mathcal{H}_0:P=Q_0$ against $\mathcal{H}_1: \mathrm{MMD}_k(P,Q_0) \geq \rho$ where $Q_0$ is a hypothesised (known) distribution, which will be specified later on.

Recall that a test is simply a function $\phi$ taking as input data and returning a binary value for whether the test rejects the null. We denote a collection of level $\alpha$ tests for one-sample testing under $r$ data corruption as 
\begin{align*}
	\Phi_{\alpha,r,Q_0} = \biggl\{\phi : \sup_{M_r \in \mathcal{M}_r}\mathbb{E}_{Y^n \sim Q_0}[\phi(M_r(Y^n))] \leq \alpha \biggr\},
\end{align*}
where $\mathcal{M}_r$ is the collection of $r$-manipulation attack functions, \emph{i.e.}, each function $M_r: \mathbb{R}^{n \times d} \mapsto \mathbb{R}^{n \times d}$ in $\mathcal{M}_r$ first chooses $r$ components of $Y^n =(Y_1,\ldots,Y_n)$ and changes them to arbitrary values on the same support. We define the class of alternatives for the one-sample problem as
\begin{align*}
	\mathcal{P}_1(\rho) \coloneqq \bigl\{ P: \mathrm{MMD}_k(P,Q_0) \geq \rho  \big\}.
\end{align*}
Choose some specific distribution $P_0 \in \mathcal{P}_1(\rho)$ and $M_r^\ast \in \mathcal{M}_r$ specified later on. By Le Cam's two point method \citep{lecam1973convergence,le2012asymptotic}, the minimax type II error satisfies 
\begin{align*}
	& \inf_{\phi \in \Phi_{\alpha,r,Q_0}} \sup_{M_r \in \mathcal{M}_r} \sup_{P \in \mathcal{P}_1(\rho)} \mathbb{E}_{Y^n \sim P}[1 - \phi(M_r(Y^n))] \\
	& \geq  \inf_{\phi \in \Phi_{\alpha,r,Q_0}}  \mathbb{E}_{Y^n \sim P_0}[1 - \phi(M_r^\ast(Y^n))]  \\
	& = 1 - \sup_{\phi \in \Phi_{\alpha,r,Q_0}}  \mathbb{E}_{Y^n \sim P_0}[\phi(M_r^\ast(Y^n))] \\
	& = 1 - \sup_{\phi \in \Phi_{\alpha,r,Q_0}}  \Bigl\{ \mathbb{E}_{Y^n \sim P_0}[\phi(M_r^\ast(Y^n))] - \mathbb{E}_{Y^n \sim Q_0}[\phi(Y^n)] + \mathbb{E}_{Y^n \sim Q_0}[\phi(Y^n)]\Bigr\} \\
	& \geq 1 - \alpha - \sup_{\phi \in \Phi_{\alpha,r,Q_0}}  \Bigl\{ \mathbb{E}_{Y^n \sim P_0}[\phi(M_r^\ast(Y^n))] - \mathbb{E}_{Y^n \sim Q_0}[\phi(Y^n)]\Bigr\}  \\
	& \geq 1 - \alpha - d_{\mathrm{TV}} \bigl(M_r^\ast(Y^n_{P_0}),  Y^n_{Q_0}  \bigr) \\
	& = 1 -  \alpha - d_{\mathrm{TV}} \bigl(M_r^\ast(Y^n_{P_0}),  F(Y^n_{P_0})  \bigr),
\end{align*}
where $F$ is an optimal transport such that $Y^n_{Q_0} \overset{d}{=} F(Y^n_{P_0})$, and
\begin{align} \label{def: coupling}
	\inf_{R \in \pi(P_0,Q_0)} \mathbb{E}_{(Y^n_{P_0},Y^n_{Q_0}) \sim R}\bigl[ d_{\mathrm{Ham}}(Y^n_{P_0},Y^n_{Q_0})) \bigr] = \mathbb{E}\bigl[ d_{\mathrm{Ham}}(Y^n_{P_0},F(Y^n_{P_0})) \bigr],
\end{align}
which is guaranteed to exist (minimiser of optimal transport).
Here $\pi(P_0,Q_0)$ denotes the set of all couplings between $P_0$ and $Q_0$. Now, following the proof of  \citet[][Theorem 1]{acharya2021robust}, take $M_r^\ast$ as 
\begin{align*}
	M_r^\ast(Y^n_{P_0}) = \begin{cases}
		F(Y^n_{P_0}), & \text{if $d_{\mathrm{Ham}}(Y^n_{P_0},F(Y^n_{P_0})) \leq r$,} \\
		Y^n_{P_0}, &  \text{if $d_{\mathrm{Ham}}(Y^n_{P_0},F(Y^n_{P_0})) > r$,}
	\end{cases}
\end{align*}
which corrupts at most $r$ samples by construction, and hence belongs to $\mathcal{M}_r$.
Hence, we get
\begin{align*}
	1 -  \alpha - d_{\mathrm{TV}} \bigl(M_r^\ast(Y^n_{P_0}),  F(Y^n_{P_0})  \bigr) & \overset{\mathrm{(i)}}{\geq} 1 -  \alpha  - \mathbb{P}\bigl( d_{\mathrm{Ham}}(Y^n_{P_0},F(Y^n_{P_0})) > r \bigr) \\
	& \overset{\mathrm{(ii)}}{\geq} 1 - \alpha - r^{-1} \mathbb{E}\bigl[ d_{\mathrm{Ham}}(Y^n_{P_0},F(Y^n_{P_0}))  \bigr] \\
	& \overset{\mathrm{(iii)}}{\geq} 1 - \alpha - r^{-1} n d_{\mathrm{TV}}(P_0,Q_0),
\end{align*}
where step~(i) follows by the coupling lemma of the TV distance, \emph{i.e.}, $d_{\mathrm{TV}}(X,Y) \leq \mP(X \neq Y)$, step~(ii) uses Markov's inequality, and step~(iii) follows by \citet[][Lemma 20]{acharya2021differentially} along with the definition of $F$ in \eqref{def: coupling}. Therefore, if 
\begin{align} \label{Eq: TV condition}
	d_{\mathrm{TV}}(P_0,Q_0) \leq \frac{r}{n} (1-\alpha-\beta),
\end{align}
then the minimax type II error is bounded below by $\beta$. Finally, as in \citet[][Appendix E.10.1]{kim2023differentially}, we choose $P_0 = p_0 \delta_x + (1-p_0) \delta_v$ and $Q_0 = q_0 \delta_x + (1-q_0) \delta_v$, where $x,v \in \mathbb{R}^d$, $0 < p_0,q_0 <1$ and $\delta_x$ is a Dirac measure at $x$, which yields
\begin{align*}
	\mathrm{MMD}_k(P_0,Q_0) = \sqrt{2\bigl(\kappa(0) - \kappa(x-v)\bigr)}\underbrace{|p_0 - q_0|}_{=d_{\mathrm{TV}}(P_0,Q_0)}.
\end{align*}  
Choose $x$ and $v$ such that $\kappa(0) - \kappa(x-v) \geq \eta$. Moreover let $q_0 = 1/2$ and $p_0 = 1/2 +r/(2n)$. Then since $\alpha + \beta < 0.4$, the condition~\eqref{Eq: TV condition} holds as
\begin{align*}
	d_{\mathrm{TV}}(P_0,Q_0) = \frac{r}{2n} \leq \frac{r}{n} (1 - \alpha - \beta)
\end{align*}
and the corresponding MMD is upper bounded as  	
\begin{align*}
\mathrm{MMD}_k(P_0,Q_0) \geq \frac{\sqrt{2\eta}}{2} \frac{r}{n}.
\end{align*}
Hence, the second part of the lower bound holds. This together with the condition $\alpha \asymp \beta$ completes the proof of \Cref{res:mmd_separation}(ii). 
\end{proof}

\subsection{Proof of \Cref{res:hsic_consistent}}

\hsicconsistent*

\begin{proof}
The proof of \Cref{res:hsic_consistent} mirrors the one of \Cref{res:mmd_consistent}. 
In order to prove the consistency of dcHSIC, it is enough to verify the condition in \Cref{res:dc_consistent}:
$$
\lim_{n \rightarrow \infty} \mP_{P_{Y\!Z}}\biggl(\,\widehat{\mathrm{HSIC}}(\widetilde{\mathcal{X}}_{n})  ~>~ \widehat{\mathrm{HSIC}}(\widetilde{\mathcal{X}}_{n}^{\bpi})+ \frac{16r(n-1)\sqrt{KL}}{n^2} \biggr) = 1.
$$
Since $r/n \to 0$, it is enough to verify that 
$$
\lim_{n \rightarrow \infty} \mP_{P_{Y\!Z}}\Bigl(\,\widehat{\mathrm{HSIC}}(\widetilde{\mathcal{X}}_{n}) >~ \widehat{\mathrm{HSIC}}(\widetilde{\mathcal{X}}_{n}^{\bpi}) \Bigr) = 1,
$$ 
which was shown in the proof of \citet[][Theorem 6]{kim2023differentially}. Therefore, dcHSIC is consistent. 
\end{proof}

\subsection{Proof of \Cref{res:hsic_separation}}

\hsicseparation*

\begin{proof}
We stress that from \Cref{res:dc_consistent}, dpHSIC controls the non-asymptotic type I error rate at level $\alpha$, so we focus on the type II error guarantees. 


(i)
The proof for the uniform separation of dcHSIC closely resembles that of dcMMD in \Cref{res:mmd_separation}. The only difference is that we need to use concentration inequalities for the empirical HSIC rather than the empirical MMD. To present details, we assume that the data are \emph{not corrupted} throughout the proof and derive the minimum uniform separation for dcHSIC. As explained in the proof of \Cref{res:mmd_separation}, this assumption does not affect the separation rate due to the inequality given in \Cref{Eq: type II error conversion}.

Denote the $(1-\alpha/2)$-quantile of the full permutation distribution as 
\begin{align*}
    q_{\infty,1-\alpha/2} = \inf\bigg\{u \in \mathbb{R} : 1 - \frac{\alpha}{2}  \leq \frac{1}{n!} \sum_{\bpi \in \boldsymbol{\Pi}_{n}} \mathds{1}\bigl( \widehat{\mathrm{HSIC}}(\widetilde{\mathcal{X}}_{n}^{\bpi}) \leq u \bigr) \bigg\},
\end{align*}
and define the event that $E_1 := \{\widetilde q \leq q_{\infty,1-\alpha/2}\}$. With $B \geq 3 \alpha^{-2}\bigl\{ \log\bigl(8/\beta\bigr) + \alpha (1-\alpha) \}$, we have $\mP_{P_{Y\!Z}}(E_1) \geq 1 - \beta/2$, which is shown in the proof of \citet[][Proposition 4]{schrab2021mmd}. Moreover, \citet[][Lemma 12]{kim2023differentially} ensures that 
\begin{align*}
    q_{\infty,1-\alpha/2} & \leq C_{1,K,L} \sqrt{\frac{1}{n} \max \biggl\{ \log\biggl( \frac{2}{\alpha}\biggr), \sqrt{\log \biggl( \frac{2}{\alpha} \biggr)}, 1\biggr\}} \leq C_{2,K,L} \sqrt{\frac{\log(e/\alpha)}{n}}
\end{align*}
where $C_{1,K,L}, C_{2,K,L}, \ldots$ are constants depending on $K$ and $L$. Define another event 
\begin{align*}
    E_2 := \biggl\{ \bigg| \mathrm{HSIC}_{k,\ell}(P_{Y\!Z}) - \widehat{\mathrm{HSIC}}(\widetilde{\mathcal{X}}_{n}) \bigg|  \leq C_{3,K,L} \sqrt{\frac{\log(e/\beta)}{n}} \biggr\}
\end{align*}
which holds with $\mP_{P_{Y\!Z}}(E_2) \geq 1 - \beta/2$ by \citet[][Lemma 14]{kim2023differentially}. With the two events $E_1$ and $E_2$, holding with high probability, the type II error of dcHSIC is bounded above as 
\begin{align*}
    & \mP_{P_{Y\!Z}} \bigl(\textrm{dcHSIC fails to reject } \mathcal{H}_0 \,\given\, \textrm{uncorrupted data} \bigr) \\
    =~ &\mP_{P_{Y\!Z}}\biggl(\widehat{\mathrm{HSIC}}(\widetilde{\mathcal{X}}_{n}) \leq \widetilde q + \frac{16r(n-1)\sqrt{KL}}{n^2} \biggr) \\
    \leq~ & \mP_{P_{Y\!Z}}\biggl( \widehat{\mathrm{HSIC}}(\widetilde{\mathcal{X}}_{n}) \leq  C_{2,K,L} \sqrt{\frac{\log(e/\alpha)}{n}} + \frac{16r\sqrt{KL}}{n} \biggr) + \mP_{P_{Y\!Z}}(E_1^c) \\
    \leq~ & \mP_{P_{Y\!Z}}\biggl( \mathrm{HSIC}_{k,\ell}(P_{Y\!Z})- C_{3,K,L} \sqrt{\frac{\log(e/\beta)}{n}} \leq C_{2,K,L} \sqrt{\frac{\log(e/\alpha)}{n}} + \frac{16r\sqrt{KL}}{n} \biggr) \\
    & \hskip 5em + \mP_{P_{Y\!Z}}(E_1^c) + \mP_{P_{Y\!Z}}(E_2^c) \\
    \leq ~ &\mP_{P_{Y\!Z}}\biggl( \mathrm{HSIC}_{k,\ell}(P_{Y\!Z}) \leq C_{4,K,L} \max \biggl\{ \sqrt{\frac{\max\{\log(e/\alpha),\log(e/\beta)\}}{n}}, \, \frac{r}{n} \biggr\}  \biggr) + \beta  = \beta,
\end{align*}
where the last equality holds by taking $C_{K,L} > C_{4,K,L}$ in the theorem statement. This proves the uniform separation of dcHSIC. 

(ii)
We would like to show that if the separation parameter $\rho$ in \Cref{def: hsic alternative} is smaller than the following threshold:
\begin{align*}
	\rho \leq C_{\eta_Y,\eta_Z} \max\biggl\{\min\biggl(\sqrt{\frac{\log(e/(\alpha+\beta))}{n}}, 1\biggr), \, \frac{r}{n} \biggr\},
\end{align*}
no test can have power greater than $1 - \beta$ uniformly over $\mathcal{P}_{\mathrm{HSIC}_{k,\ell}}(\rho)$ where $C_{\eta_Y,\eta_Z} > 0$ is a small constant depending on $\eta_Y,\eta_Z$ in \Cref{assumption: independence kernel}. The first part of the lower bound, involving $n^{-1/2}$, was obtained in \citet[][Appendix F.4.1]{kim2023differentially} with a slight modification to a constant factor $C_{\eta_Y,\eta_Z}$ to replace $\log(1/(\alpha+\beta))$ with $\log(e/(\alpha+\beta))$ where $\alpha + \beta < 0.4$. Hence, it is enough to prove the second part of the threshold, involving $r/n$. The proof of this claim is essentially the same as the proof of \Cref{res:mmd_separation}(ii).
To prove the desired result, it hence suffices to find a distribution $P_{Y\!Z}$ such that 
\begin{equation}
\begin{aligned} \label{Eq: condition for HSIC lower bound}
	& d_{\mathrm{TV}}(P_{Y\!Z}, P_Y \times P_Z) \leq \frac{r}{n} (1- \alpha -\beta) \quad \text{and} \\
	& \mathrm{HSIC}_{k,\ell}(P_{Y\!Z}) \geq C_{\eta_Y,\eta_Z} \frac{r}{n}.
\end{aligned}
\end{equation}
To this end, as in \citet[][Appendix F.4.1]{kim2023differentially}, construct the distribution of $(Y,Z)$ as
\begin{align*}
	& \mP(Y = y_1, Z = z_1) = \mP(Y = y_2, Z = z_2) = 1/4 + \mu \quad \text{and} \\
	& \mP(Y = y_1, Z = z_2) = \mP(Y = y_2, Z = z_1) = 1/4 - \mu, 
\end{align*}
for some distinct $y_1,y_2 \in \mathbb{R}^{d_Y}$ and $z_1,z_2 \in \mathbb{R}^{d_Z}$ such that $y_1-y_2 = y_0$ and $z_1-z_2 = z_0$, and $\mu \in (0,1/4]$. Following the calculations in \citet[][Appendix F.4.1]{kim2023differentially}, the HSIC of such $P_{Y\!Z}$ and the TV distance between $P_{Y\!Z}$ and $P_Y \times P_Z$ can be computed as
\begin{align*}
	 \mathrm{HSIC}_{k,\ell}(P_{Y\!Z}) \geq  2 \mu \sqrt{\eta_Y\eta_Z} \quad \text{and}  \quad  d_{\mathrm{TV}}(P_{Y\!Z}, P_Y \times P_Z) = 2\mu. 
\end{align*}
Therefore, given that $\alpha + \beta < 0.4$, the condition~\eqref{Eq: condition for HSIC lower bound} is fulfilled by setting $\mu = r/(4n)$. This, together with the condition $\alpha \asymp \beta$, completes the proof of \Cref{res:hsic_separation}(ii). 
\end{proof}

\section{ROBUSTNESS PROPERTIES OF THE RELATED DP PROCEDURE}
\label{sec:dp_results}

In this section, we derive results on the robustness properties of the DP tests adapted from \citet{kim2023differentially} and presented in \Cref{sec:related} and \Cref{alg:dp}.

First, we guarantee the validy of DP tests under robustness, and provide a condition guaranteeing their consistency in the data corruption framework. 

\begin{restatable}[Validity \& consistency of DP]{lemma}{dpconsistent}
\label{apres:dp_consistent_ap}
 (i) The DP test of \Cref{alg:dp} has non-asymptotic level control under $r$ data corruption.
(ii) Let $P_1\in\mathcal{P}_1$ be an alternative distribution.
Assume $\alpha\in(0,1)$ fixed and $r\varepsilon \leq \nu$ for all $n \in \mathbb{N}$ where $\nu$ is some positive constant. For any sequence of $B_n$ of permutation numbers satisfying $\min_{n\in\mathbb{N}} B_n > e^\nu \alpha^{-1} - 1$, the DP test is consistent in the sense that $\lim_{n \rightarrow \infty} \mP_{P_1}\bigl(\textrm{\emph{DP rejects }} \mathcal{H}_0 \given r \textrm{\emph{ corrupted data}} \bigr) = 1$ if
$$
\lim_{n \rightarrow \infty} \mP_{P_1}\left(\!\,T(\tXn) + \frac{2 \zeta\Delta_T }{\varepsilon} >~ T(\tXn^{\bpi}) + \frac{2 \zeta'\Delta_T }{\varepsilon} \!\right) = 1
$$
where the probability is taken with respect to the (uniformly) random permutation $\bpi$ of $[n]$, to $\tXn$ i.i.d.~drawn from $P_1$, and to $\zeta$ and $\zeta'$ i.i.d. $\mathsf{Laplace}(0,1)$ noise.
\end{restatable}



\begin{proof}
(i) The non-asympotic validity of the adjusted DP test in the data corruption framework follows from \Cref{eq:valid_dp} since the test using the uncorrupted data is non-asymptotically valid (in the usual framework) and differentially private \citep[Theorems 1 and 2]{kim2023differentially}.

\medskip
(ii) By the DP group privacy property~\citep[Theorem 2.2]{dwork2014algorithmic}, we get the type II error bounded above as
\begin{align*}
	\mP_{P_1} \bigl( \textrm{DP fails to reject $\mathcal{H}_0$} \,\given\, \textrm{corrupted data} \bigr) & \leq~ e^{r\varepsilon} \mP_{P_1} \bigl( \textrm{DP fails to reject $\mathcal{H}_0$} \,\given\, \textrm{uncorrupted data} \bigr) \\
	&\leq~ e^\nu \mP_{P_1} \bigl( \textrm{DP fails to reject $\mathcal{H}_0$} \,\given\, \textrm{uncorrupted data} \bigr)
\end{align*}
where the second inequality uses the condition $r \varepsilon \leq \nu$. Moreover, since the $(1-\alpha e^{-r\varepsilon})$-quantile of $M_0,\ldots,M_B$ is smaller than or equal to the $(1-\alpha e^{-\nu})$-quantile of $M_0,\ldots,M_B$, say $q_{1 - \alpha e^{-\nu}}$, we have
\begin{align*}
    \mP_{P_1} \bigl( \textrm{DP fails to reject $\mathcal{H}_0$} \,\given\, \textrm{uncorrupted data} \bigr) & =  \mP_{P_1} \bigl( T_0 \leq q \,\given\, \textrm{uncorrupted data} \bigr) \\
    & \leq \mP_{P_1} \bigl( T_0 \leq q_{1 - \alpha e^{-\nu}} \,\given\, \textrm{uncorrupted data} \bigr).
\end{align*}
By \citet[][Theorem 3]{kim2023differentially}, the above bound converges to zero as $n \rightarrow \infty$ under the conditions of \Cref{apres:dp_consistent_ap}. This proves that the DP test is consistent in power. 
\end{proof}

Using the above result, we can now show the consistency of the dpMMD and dpHSIC tests under data corruption.

\begin{restatable}[Consistency of dpMMD]{lemma}{mmdconsistentap} 
	\label{apres:mmd_consistent_ap}
	Suppose that the kernel $k$ is characteristic, non-negative, and bounded everywhere by $K$. 
	Assume that $n \leq m$, $r\varepsilon \leq \nu$ for some positive constant $\nu$ and 
    $1/(\varepsilon n) \to 0$
    as $n\to\infty$, and that a sequence of permutation numbers $B_n$ satisfies $\min_{n\in\mathbb{N}} B_n > e^\nu\alpha^{-1} - 1$. Then, for any fixed $P$ and $Q$ with $P \neq Q$, dpMMD is consistent in the sense that 
	$$
	\lim_{n \rightarrow \infty} \mP_{P,Q}\bigl(\textrm{\emph{reject $\mathcal{H}_0$}} \given r \textrm{\emph{ corrupted data}} \bigr) = 1.
	$$
\end{restatable}



\begin{proof}
We need to verify the condition in \Cref{apres:dp_consistent_ap} that 
$$
\lim_{n \rightarrow \infty} \mP_{P,Q}\biggl(\,\widehat{\mathrm{MMD}}(\widetilde{\mathcal{X}}_{n+m}) + 2 \zeta \frac{\sqrt{2K}}{n} \varepsilon^{-1} >~ \widehat{\mathrm{MMD}}(\widetilde{\mathcal{X}}_{n+m}^{\bpi}) + 2 \zeta' \frac{\sqrt{2K}}{n} \varepsilon^{-1} \biggr) = 1.
$$
Indeed, since $1/(\varepsilon n) \to 0$, it suffices to verify that 
$$
\lim_{n \rightarrow \infty} \mP_{P,Q}\Bigl(\,\widehat{\mathrm{MMD}}(\widetilde{\mathcal{X}}_{n+m}) >~ \widehat{\mathrm{MMD}}(\widetilde{\mathcal{X}}_{n+m}^{\bpi}) \Bigr) = 1,
$$ 
which was shown in the proof of \citet[][Theorem 5]{kim2023differentially}. Therefore, dpMMD is consistent. 
\end{proof}

\begin{restatable}[Consistency of dpHSIC]{lemma}{hsicconsistentap} 
	\label{apres:hsic_consistent_ap}
	Suppose that the kernels $k$ and $\ell$ are characteristic, non-negative, and bounded everywhere by $K$ and $L$, respectively. 
	Assume that $r\varepsilon \leq \nu$ for some positive constant $\nu$ and 
    $1/(\varepsilon n) \to 0$ 
    as $n \to \infty$, and that a sequence of permutation numbers $B_n$ satisfies $\min_{n\in\mathbb{N}} B_n > e^\nu\alpha^{-1} - 1$. Then, for any fixed joint distribution $P_{Y\!Z}$ with $P_{Y\!Z}\neq P_Y \times P_Z$, dpHSIC is consistent in the sense that 
	$$
	\lim_{n \rightarrow \infty} \mP_{P_{Y\!Z}}\bigl(\textrm{\emph{reject $\mathcal{H}_0$}} \given r \textrm{\emph{ corrupted data}} \bigr) = 1.
	$$
\end{restatable}



\begin{proof}
The proof of \Cref{apres:hsic_consistent_ap} follows a similar approach to that of \Cref{apres:mmd_consistent_ap}. 
By \Cref{apres:dp_consistent_ap}, the consistency of dpHSIC holds provided that
$$
\lim_{n \rightarrow \infty} \mP_{P_{Y\!Z}}\biggl(\,\widehat{\mathrm{HSIC}}(\widetilde{\mathcal{X}}_{n}) + 2 \zeta \frac{4(n-1)\sqrt{KL}}{n^2} \varepsilon^{-1} >~ \widehat{\mathrm{HSIC}}(\widetilde{\mathcal{X}}_{n}^{\bpi}) + 2 \zeta' \frac{4(n-1)\sqrt{KL}}{n^2} \varepsilon^{-1} \biggr) = 1.
$$
Since $1/(\varepsilon n) \to 0$, it suffices to verify that 
$$
\lim_{n \rightarrow \infty} \mP_{P_{Y\!Z}}\Bigl(\,\widehat{\mathrm{HSIC}}(\widetilde{\mathcal{X}}_{n}) >~ \widehat{\mathrm{HSIC}}(\widetilde{\mathcal{X}}_{n}^{\bpi}) \Bigr) = 1,
$$ 
which was shown in the proof of \citet[][Theorem 6]{kim2023differentially}. Therefore, dpHSIC is consistent. 
\end{proof}

Finally, we can derive uniform separation rates for the robust DP tests in terms of MMD and HSIC separation.

\begin{restatable}[Uniform separation of dpMMD]{theorem}{mmdseparationap}  
	\label{apres:mmd_separation_ap}
	Suppose that the kernel $k$ is characteristic, non-negative, and bounded everywhere by $K$. For $\alpha,\beta \in (0,1)$, assume that the number of permutations is greater than $6\alpha_{\mathtt{dp}}^{-1} \log(2/\beta_{\mathtt{dp}})$ where $\alpha_{\mathtt{dp}} = e^{-r\varepsilon} \alpha$ and $\beta_{\mathtt{dp}} = e^{-r\varepsilon} \beta$, 
setting $\varepsilon = r^{-1}\max\{\log(e/\alpha),\log(e/\beta)\}$, 
and $n \asymp m$. 
The dpMMD test is guaranteed to have high power, i.e.,
		$\mP_{P,Q}\bigl(\textrm{\emph{reject $\mathcal{H}_0$}} \given r \textrm{\emph{ corrupted data}} \bigr) \geq 1 -\beta$
		for any distributions $P$ and $Q$ separated as
		\begin{align*}
			&\mathrm{MMD}_k(P,Q)
			~\geq~
			C_{K} \max \biggl\{ \sqrt{\frac{\max\{\log(e/\alpha),\log(e/\beta)\}}{n}}, \ \frac{r}{n} \biggr\} 
		\end{align*}
		for some positive constant $C_{K}$ depending on $K$.
\end{restatable}


\begin{proof}
We stress that from \Cref{apres:dp_consistent_ap}, the dpMMD test has non-asymptotic level $\alpha$. Therefore, we focus on controlling the type II error. 
	Recall that the dpMMD test with level $\alpha$ is defined with privacy parameters $\varepsilon = r^{-1} \max\!\big\{\!\log(e/\alpha), \, \log(e/\beta)\big\}$ and $\delta=0$, with adjusted level parameter $\alpha_{\mathtt{dp}} \coloneqq e^{-r\varepsilon} \alpha$. As in the proof of \Cref{apres:dp_consistent_ap}, we may use the DP group property to bound the type II error of dpMMD as
\begin{align*}
	&\mP_{P,Q} \bigl( \textrm{dpMMD fails to reject $\mathcal{H}_0$} \,\given\, \textrm{corrupted data} \bigr) \\
    \leq~ &e^{r\varepsilon} \mP_{P,Q} \bigl( \textrm{dpMMD fails to reject $\mathcal{H}_0$} \,\given\, \textrm{uncorrupted data} \bigr) \\
	\leq ~ &e^{r\varepsilon} \beta_{\mathtt{dp}} \coloneqq \beta.
\end{align*}
Let us define the class of pairs of distributions, which are $\rho$-separated in terms of the MMD metric, as
\begin{align} \label{def: MMD alternative}
	\mathcal{P}_{\mathrm{MMD}_k}(\rho) = \bigl\{ (P,Q) : \mathrm{MMD}_k(P,Q) \geq \rho \bigr\}.
\end{align}
Now, by leveraging \citet[][Theorem 7]{kim2023differentially} with $n \asymp m$ and $B \geq 6\alpha_{\mathtt{dp}}^{-1}\log(2\beta_{\mathtt{dp}}^{-1})$, we see that the minimum value of $\rho$ that controls the type II error as
\begin{align*}
	\sup_{(P,Q) \in \mathcal{P}_{\mathrm{MMD}_k}(\rho)}\mP_{P,Q} \bigl( \textrm{dpMMD fails to reject $\mathcal{H}_0$} \,\given\, \textrm{uncorrupted data} \bigr) ~\leq~ \beta_{\mathtt{dp}},
\end{align*}
satisfies
\begin{align*}
	 \rho ~\leq~ &C_{K}' \max \Biggl\{ \sqrt{\frac{\max\!\big\{\!\log(e/\alpha_{\mathtt{dp}}), \, \log(e/\beta_{\mathtt{dp}})\big\}}{n}}, \, \frac{\max\!\big\{\!\log(e/\alpha_{\mathtt{dp}}), \, \log(e/\beta_{\mathtt{dp}})\big\}}{n \varepsilon}  \Biggr\}\\
	\leq~ &C_{K}' \max \Biggl\{ \sqrt{\frac{r\varepsilon + \max\!\big\{\!\log(e/\alpha), \, \log(e/\beta)\big\}}{n}}, \, \frac{r\varepsilon + \max\!\big\{\!\log(e/\alpha), \, \log(e/\beta)\big\}}{n \varepsilon}  \Biggr\}\\
	\leq~ &C_{K} \max \Biggl\{ \sqrt{\frac{r\varepsilon}{n}},\, \frac{r}{n},\, \sqrt{\frac{\max\!\big\{\!\log(e/\alpha), \, \log(e/\beta)\big\}}{n}}, \, \frac{\max\!\big\{\!\log(e/\alpha), \, \log(e/\beta)\big\}}{n \varepsilon}  \Biggr\} \\
	= ~ & C_{K} \max \Biggl\{\sqrt{\frac{\max\!\big\{\!\log(e/\alpha), \, \log(e/\beta)\big\}}{n}},\frac{r}{n} \Biggr\},
\end{align*}
where $C_{K}', C_{K}$ are constants depending on $K$.
The last equality holds since we set $\varepsilon = r^{-1} \max\!\big\{\!\log(e/\alpha), \, \log(e/\beta)\big\}$. This proves the uniform separation of the dpMMD test. 
\end{proof}

\begin{restatable}[Uniform separation of dpHSIC]{theorem}{hsicseparationap} 
	\label{apres:hsic_separation_ap}
	Suppose that the kernels $k$ and $\ell$ are characteristic, non-negative, and bounded everywhere by $K$ and $L$, respectively. For $\alpha,\beta \in (0,1)$, assume that the number of permutations is greater than $6\alpha_{\mathtt{dp}}^{-1} \log(2/\beta_{\mathtt{dp}})$ where $\alpha_{\mathtt{dp}} = e^{-r\varepsilon} \alpha$ and $\beta_{\mathtt{dp}} = e^{-r\varepsilon} \beta$, 
setting $\varepsilon = r^{-1}\max\{\log(e/\alpha),\log(e/\beta)\}$.
The dpHSIC test is guaranteed to have high power, i.e.,
		$\mP_{P_{Y\!Z}}\bigl(\textrm{\emph{reject $\mathcal{H}_0$}} \given r \textrm{\emph{ corrupted data}} \bigr) \geq 1 -\beta$ for any joint distribution $P_{Y\!Z}$ separated as
		\begin{align*}
			&\mathrm{HSIC}_{k,\ell}(P_{Y\!Z})
			~\geq~C_{K,L} \max \biggl\{ \sqrt{\frac{\max\{\log(e/\alpha),\log(e/\beta)\}}{n}}, \ \frac{r}{n} \biggr\} 
		\end{align*}
		for some positive constant $C_{K,L}$ depending on $K$ and $L$. 
\end{restatable}



\begin{proof}
We emphasize that from \Cref{apres:dp_consistent_ap}, the dpHSIC test has non-asymptotic level $\alpha$. Therefore, we focus on controlling the type II error.
Recall that the dpHSIC test with level $\alpha$ is defined with privacy parameters $\varepsilon = r^{-1} \max\!\big\{\!\log(e/\alpha), \, \log(e/\beta)\big\}$ and $\delta=0$, with adjusted level parameter $\alpha_{\mathtt{dp}} \coloneqq e^{-r\varepsilon} \alpha$. As in the proof of \Cref{apres:dp_consistent_ap}, we may use the DP group property to bound the type II error of dpHSIC as
\begin{align*}
	&\mP_{P_{Y\!Z}} \bigl( \textrm{dpHSIC fails to reject $\mathcal{H}_0$} \,\given\, \textrm{corrupted data} \bigr) \\
    \leq~ &e^{r\varepsilon} \mP_{P_{Y\!Z}} \bigl( \textrm{dpHSIC fails to reject $\mathcal{H}_0$} \,\given\, \textrm{uncorrupted data} \bigr) \\
	\leq ~ &e^{r\varepsilon} \beta_{\mathtt{dp}} \coloneqq \beta.
\end{align*}
Let us define the class of distributions, which are $\rho$-separated in terms of the HSIC metric, as
\begin{align} \label{def: hsic alternative}
	\mathcal{P}_{\mathrm{HSIC}_{k,\ell}}(\rho) = \bigl\{ P_{Y\!Z} : \mathrm{HSIC}_{k,\ell}(P_{Y\!Z}) \geq \rho \bigr\}.
\end{align}
Now, by leveraging \citet[][Theorem 12]{kim2023differentially} and $B \geq 6\alpha_{\mathtt{dp}}^{-1}\log(2\beta_{\mathtt{dp}}^{-1})$, we see that the minimum value of $\rho$ that controls the type II error as
\begin{align*}
	\sup_{P_{Y\!Z} \in \mathcal{P}_{\mathrm{HSIC}_{k,\ell}}(\rho)}\mP_{P_{Y\!Z}} \bigl( \textrm{dpHSIC fails to reject $\mathcal{H}_0$} \,\given\, \textrm{uncorrupted data} \bigr) \leq \beta_{\mathtt{dp}},
\end{align*}
satisfies
\begin{align*}
	\rho ~\leq~ &C_{K,L}' \max \Biggl\{ \sqrt{\frac{\max\!\big\{\!\log(e/\alpha_{\mathtt{dp}}), \, \log(e/\beta_{\mathtt{dp}})\big\}}{n}}, \, \frac{\max\!\big\{\!\log(e/\alpha_{\mathtt{dp}}), \, \log(e/\beta_{\mathtt{dp}})\big\}}{n \varepsilon}  \Biggr\}\\
	\leq~ &C_{K,L}' \max \Biggl\{ \sqrt{\frac{r\varepsilon + \max\!\big\{\!\log(e/\alpha), \, \log(e/\beta)\big\}}{n}}, \, \frac{r\varepsilon + \max\!\big\{\!\log(e/\alpha), \, \log(e/\beta)\big\}}{n \varepsilon}  \Biggr\}\\
	\leq~ &C_{K,L} \max \Biggl\{ \sqrt{\frac{r\varepsilon}{n}},\, \frac{r}{n},\, \sqrt{\frac{\max\!\big\{\!\log(e/\alpha), \, \log(e/\beta)\big\}}{n}}, \, \frac{\max\!\big\{\!\log(e/\alpha), \, \log(e/\beta)\big\}}{n \varepsilon}  \Biggr\} \\
	= ~ & C_{K,L} \max \Biggl\{\sqrt{\frac{\max\!\big\{\!\log(e/\alpha), \, \log(e/\beta)\big\}}{n}},\frac{r}{n} \Biggr\},
\end{align*}
where $C_{K,L}', C_{K,L}$ are constants depending on $K$ and $L$, and where the last equality holds since we set $\varepsilon = r^{-1} \max\!\big\{\!\log(e/\alpha), \, \log(e/\beta)\big\}$. This proves the uniform separation of the dpHSIC test. 
\end{proof}

\end{document}